\documentclass[10pt,twocolumn,letterpaper]{article}

\usepackage{cvpr}              

\usepackage{graphicx}
\usepackage{amsmath}
\usepackage{amssymb}
\usepackage{booktabs}
\usepackage{multicol}
\usepackage{multirow}
\usepackage{comment}
\usepackage{mathtools, nccmath}
\usepackage{pifont}
\newcommand{\cmark}{\ding{51}}%
\newcommand{\xmark}{\ding{55}}%
\usepackage{hyperref}
\hypersetup{breaklinks,colorlinks}
\usepackage[capitalize]{cleveref}
\crefname{section}{Sec.}{Secs.}
\Crefname{section}{Section}{Sections}
\Crefname{table}{Table}{Tables}
\crefname{table}{Tab.}{Tabs.}

\def\cvprPaperID{4626} 
\def\confName{CVPR}
\def\confYear{2022}

\begin{document}

\title{Human Instance Matting via Mutual Guidance and Multi-Instance Refinement}

\author{Yanan Sun$^{1,2}$~~~~~~~~~~~~~~~~~~
Chi-Keung Tang$^{1}$~~~~~~~~~~~~~~~~~~
Yu-Wing Tai$^{2}$\\
\vspace{1mm}
$^{1}$HKUST~~~~~~~~~~~~~~~~~~$^{2}$Kuaishou Technology\\
{\tt\small 
\{now.syn, yuwing\}@gmail.com~~~~~~~ 
cktang@cs.ust.hk}
}
\maketitle

\begin{abstract}
This paper introduces a new matting task called human instance matting (HIM), which requires the pertinent model to automatically predict a precise alpha matte for each human instance.  Straightforward combination of  closely related techniques, namely, instance segmentation, soft segmentation and human/conventional matting, will easily fail in complex cases requiring disentangling mingled colors belonging to multiple instances along hairy and thin boundary structures.  To tackle these technical challenges, we propose a human instance matting framework, called InstMatt, where a novel mutual guidance strategy working in tandem with a multi-instance refinement module is used, for delineating multi-instance relationship among humans with complex and overlapping boundaries if present. A new instance matting metric called instance matting quality (IMQ) is proposed, which addresses the absence of a unified and fair means of evaluation emphasizing  both instance recognition and matting quality. Finally, we construct a HIM benchmark for evaluation, which comprises of both synthetic and natural benchmark images. In addition to thorough experimental results on complex cases with multiple and overlapping human instances each has intricate boundaries, preliminary results are presented on general instance matting. Code and benchmark are available in 
{\small\url{https://github.com/nowsyn/InstMatt}}. 
{\let\thefootnote\relax\footnotetext{{This work was done when Yanan Sun was a student intern at Kuaishou Technology, which was supported by Kuaishou Technology and the Research Grant Council of the Hong Kong SAR under grant no. 16201420.}}}
\end{abstract}

\vspace{-0.2in}
\section{Introduction}
\begin{figure}[t]
    \centering
    \includegraphics[width=1.0\linewidth]{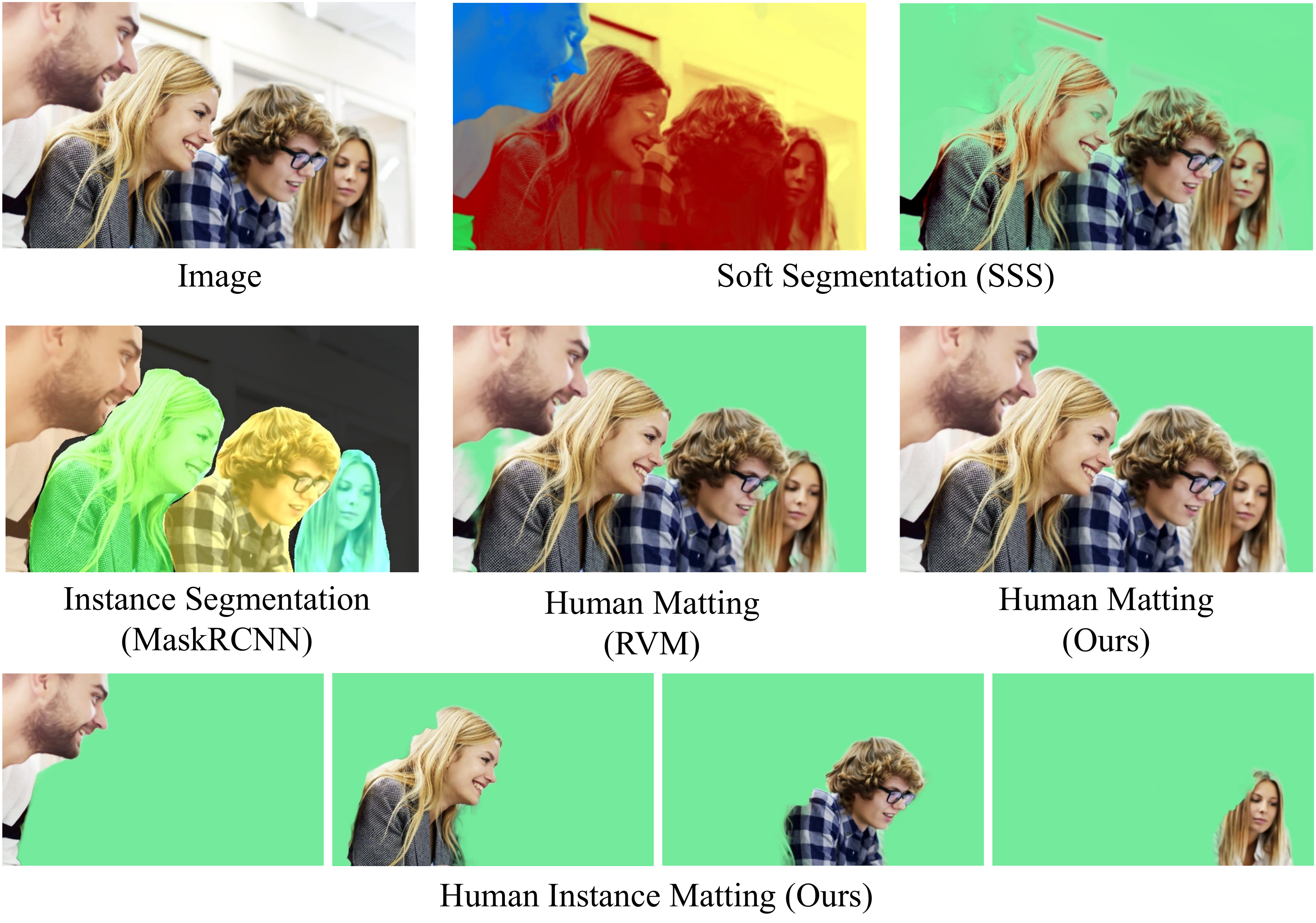}
    \vspace{-0.25in}
    \caption{Comparisons with related tasks, including soft segmentation results from SSS~\cite{SSS}, instance segmentation results from MaskRCNN~\cite{MaskRCNN}, human matting results from RVM~\cite{rvm} as well as ours, and human instance matting results from ours.}
    \label{fig:teaser}
    \vspace{-0.3in}
\end{figure}

Fast development of mobile internet technology has triggered the rapid growth of multimedia industry especially we-media, where users are heavily engaged  in editing tools to beautify or re-create their image and video contents. As one of the primary techniques for efficient image editing, image matting has achieved significant improvement with the wide adoption of deep neural networks in the task. However, existing matting methods still fail or else are not easy to use in  many scenarios, such as extracting the foreground human while removing background humans, or instance-level editing as shown in Figure~\ref{fig:teaser}: what if we want to independently extract and edit each human instance? 

Similar to semantic versus instance segmentation, existing matting methods, which focus on a region based on a given trimap or a known object class, are unable to differentiate instances. 
To address this issue,  we propose a new task called human instance matting (HIM), which aims to automatically extract precise alpha matte for each human instance in a given image. HIM shares similarities to the following conventional tasks while embodying fundamental  differences making it a problem on its own:  
1) instance segmentation aims at distinguishing instances, but it can only produce sharp object boundary without semi-transparency consideration; 2) recent soft segmentation~\cite{SSS} is capable of generating soft segments for multiple instances of different classes with instance-aware features, but cannot deal with instances of the same class; 3) conventional matting aims at extracting precise alpha matte, but it lacks instance awareness. Overall, human instance matting is a unified task encompassing the characteristics of the aforementioned related tasks while introducing new technical challenges. 

Conventional matting is based on the image compositing equation where an image $I$ is the combination of foreground $F$ layer, background $B$ layer modulated by alpha $\alpha$:
\begin{equation}\label{eq:alpha}
I = \alpha F + (1-\alpha) B.
\end{equation}
To adapt to multiple instance matting, we modify the 2-layer Equation~\ref{eq:alpha} to one of multi-instance layered composition, where each instance layer is attenuated by its corresponding $\alpha$: 
\begin{equation}\label{eq:im}
I = \sum_{i=0}^{n}\alpha_iL_i, \hspace{0.2in} \textrm{s.t.}   \sum_{i=0}^{n}\alpha_i = 1
\end{equation}
where $L_i$ and $\alpha_i$ respectively denote the foreground and alpha matte for instance $i>0$; $L_0$ and $\alpha_0$ respectively represent the background and its corresponding alpha matte; $n$ is the number of instances.
This equation had also appeared in~\cite{spectral, SSS}, but all such relevant matting and segmentation tasks were not instance aware. 
The goal of instance  matting is to solve for target mattes $\alpha_i$ for all $i>0$.

By exploring the complex relation among multiple instances, we propose a new instance matting framework, called \textbf{InstMatt}, where a novel {\bf mutual guidance strategy} enables a deep model to decompose mingled compositing colors into their respective instances. Our mutual guidance strategy takes both the relation between instances and the background, and the relation among instances into consideration. Besides, a {\bf multi-instance refinement} module is carefully designed and engineered for interchanging information among instances to synchronize predictions for further refinement. Equipped with the novel mutual guidance and multi-instance refinement, our InstMatt is able to not only produce high-quality human alpha matte but also distinguish multiple human instances shown in Figure~\ref{fig:teaser}.

With this new HIM task, existing evaluation metrics  for  instance segmentation or matting are insufficient, which were designed for either one of the tasks. 
We propose a new metric, called instance matting quality ({\bf IMQ}), that  simultaneously measures instance recognition quality and alpha matte quality. To provide a general and comprehensive validation on instance matting techniques, we construct an instance matting benchmark, {\bf HIM2K}, which consists of a synthetic image benchmark and a natural image benchmark totaling 2,000 images with high-quality matte ground truths. 

To demonstrate the promise of our technical contributions beyond human instance matting, we present preliminary results on matting multi-object instances not limited to humans, a fruitful future direction to explore.

\section{Related Work}
\subsection{Matting} 
\vspace{-2mm}
\noindent{\textbf{Natural Image Matting.}}
Image matting is a pixel-level task, aiming to extract alpha matte for a foreground object. Traditional matting methods can be summarized into two approaches. Sampling-based methods~\cite{Bayesian, globalsampling, sharedsampling, globalsampling} collect a set of known foreground and background samples to estimate unknown alpha values. Propagation-based methods~\cite{spectral, knnmatting, closedform, randomwalks, geodesic, ifm} assume  neighboring pixels are correlated, and use their affinities to propagate alpha from known regions to unknown regions. Traditional methods rely on low-level or statistical features, which can easily fail on complex cases due to their limited feature representation.

The wide application of deep convolutional neural network (CNN) addresses this feature representation issue to a great extent. DCNN~\cite{DCNN} and DIM~\cite{DIM} are the first representative methods to apply CNN in matting, which are followed by a series of valuable works advancing the state-of-the-art matting performance. Deep learning-based methods can be further grouped into three approaches. Trimap (or mask) based methods~\cite{AlphaGAN, IndexNet, adamatting, ContextAware, ISBM, GCA, Affinity, SIM, DVM, MG, FBA, videomatting_2021_ICCV, imagematting_2021_ICCV} take an additional trimap to focus the model on the target foreground object. With careful network design, these methods have achieved excellent performance. 
User-supplied constraints 
are relaxed in~\cite{BGM, BGMv2} by using an extra photo taken without the relevant foreground object for providing useful prior information. Trimap-free methods~\cite{LFM, HAttNet} erase the dependence on  additional input. These methods resort attention or salience to localize foreground object and extract the corresponding alpha matte.

\vspace{0.1in}
\noindent{\textbf{Human Matting.}}
Human matting is a class-specific image matting task, where the semantic information of the foreground object, namely, human is known. Known human semantics effectively guides relevant human matting methods and thus they usually do not require additional input. Deep learning-based human matting was first proposed in~\cite{deep_portrait} and then improved in SHM~\cite{SHM}. A method was proposed in BSHM~\cite{BSHM} which makes use of coarse annotated data for boosting performance. MODNet~\cite{MODNet} addresses automatic and fast human matting using a light-weight network considering both low-resolution semantics and high-resolution details. In RVM~\cite{rvm} a video human matting framework was proposed using a recurrent decoder to improve robustness. Further, a cascade framework is proposed in~\cite{humanmatting_2021_ICCV} to extract alpha matte from low-to-high resolution.

\begin{figure*}[t]
    \centering
    \includegraphics[width=1.0\linewidth]{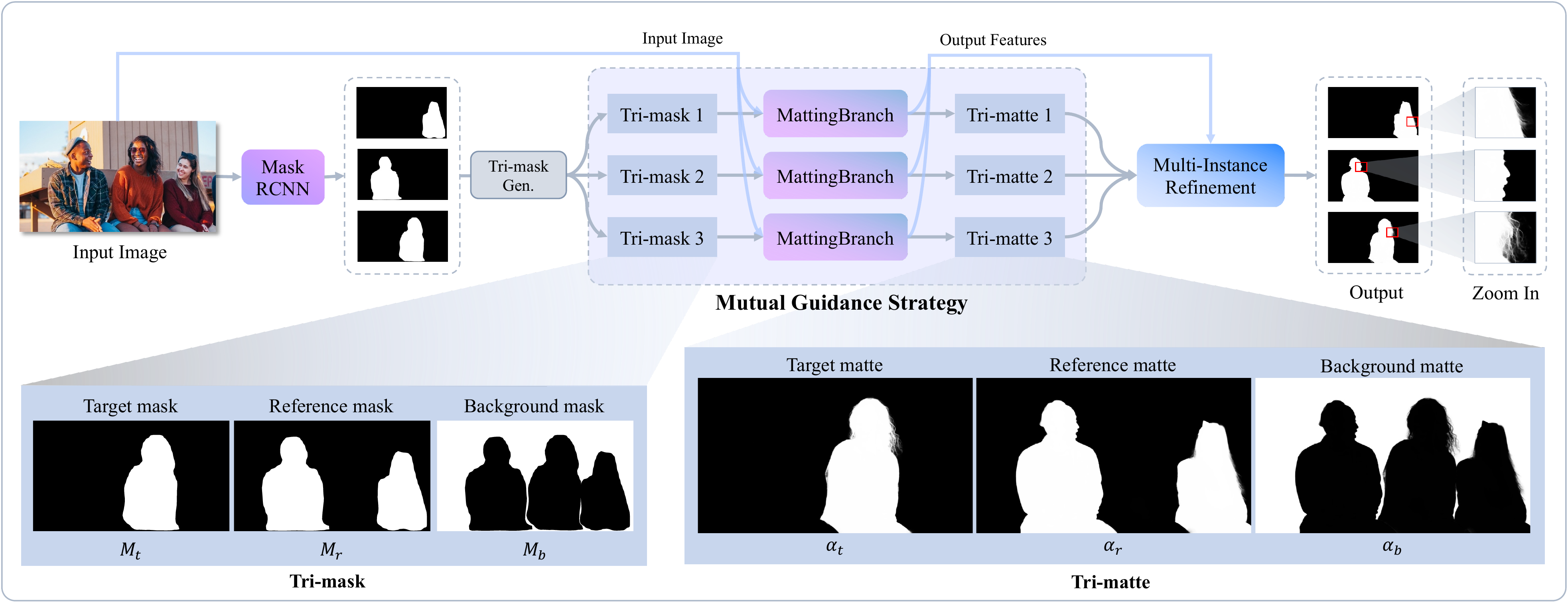}
    \vspace{-0.25in}
    \caption{Overall \textbf{InstMatt} framework consisting of mutual guidance and multi-instance refinement. We first apply MaskRCNN to obtain instance masks, and then generate \textbf{tri-mask} for each instance to provide \textbf{mutual guidance} for the matting branch. Through mutual guidance strategy, we upgrade coarse tri-masks into fine \textbf{tri-mattes} for all instances. Finally, a \textbf{multi-instance refinement} module (illustrated in Figure~\ref{fig:CRH}) is designed to make use of the information difference of underlining tri-mattes to further promote the instance matte quality.}
    \label{fig:framework}
    \vspace{-0.2in}
\end{figure*}

\vspace{-0.1in}
\subsection{Segmentation}
\vspace{-2mm}
\noindent{\textbf{Instance Segmentation.}} Instance segmentation simultaneously requires instance-level and pixel-level predictions. The existing methods can be classified into three categories. Top-down methods~\cite{MaskRCNN, FCIS, PANet, MaskScore, TensorMask, YOLACT, BlendMask, bcnet} first detect instances and then segment the object within detected bounding boxes. On the contrary, bottom-up methods~\cite{SGN, SSAP} first learns the embeddings for each pixel and then group them into instances. Direct methods~\cite{SOLO, SOLOv2} are box-free and grouping-free. They predict instance masks with classification in one shot without a detection or clustering step.

\vspace{0.1in}
\noindent{\textbf{Soft Segmentation.}}
Soft segmentation is a pixel-level task, decomposing an image
into several segments where each pixel may belong partially
to multiple segments. Different decomposition methods lead to different segments. For instance, soft color segmentation methods~\cite{MIL, SCS, DIL, ColorMix, Unmix} decompose an image into soft layers of homogeneous colors; spectral matting~\cite{spectral} clusters an image into a set of spectral segments; SSS~\cite{SSS} decomposes an image into soft semantic segments via aggregating high-level embeddings with local-level textures.

\vspace{-0.06in}
\subsection{Instance Matting}
\vspace{-0.05in}
Instance matting maps each pixel into a set of soft or fractional alphas each tagged with an unique instance ID. Besides inheriting the difficulties from instance segmentation and soft segmentation, instance matting introduces new algorithmic challenges. Specifically, compared to instance segmentation, each pixel in instance matting can partially belong to more than one instance; compared to soft segmentation, each pixel can belong to multiple instances of the same class. To the best of our knowledge, there is no unified framework that can simultaneously address these technical challenges brought by the new instance matting problem. 
In this paper, we take human instance matting as an example, and propose a framework to address the aforementioned issues via our novel mutual guidance and multi-instance refinement.

\vspace{-0.05in}
\section{Method}
\vspace{-0.05in}
Our HIM framework, called \textbf{InstMatt}, consists of two steps, 
first recognizing instances and then extracting their respective alpha mattes. This allows the model to globally discover instances and then refine them according to local context. Figure~\ref{fig:framework} illustrates the whole framework.

\subsection{Observations}
\vspace{-1mm}
\paragraph{Sparsity.}
Equation~\ref{eq:im}
indicates that  a given pixel can belong to multiple instances and hence $\alpha_i, i=1\cdots n$. However, in real-life images even containing many instances, each pixel usually consists of no more than two non-zero $\alpha$s, belonging to an instance and the background, or two overlapping instances, thus satisfying the sparsity observation of multi-instance matting. 

\begin{figure*}[t]
    \centering
    \includegraphics[width=1.0\linewidth]{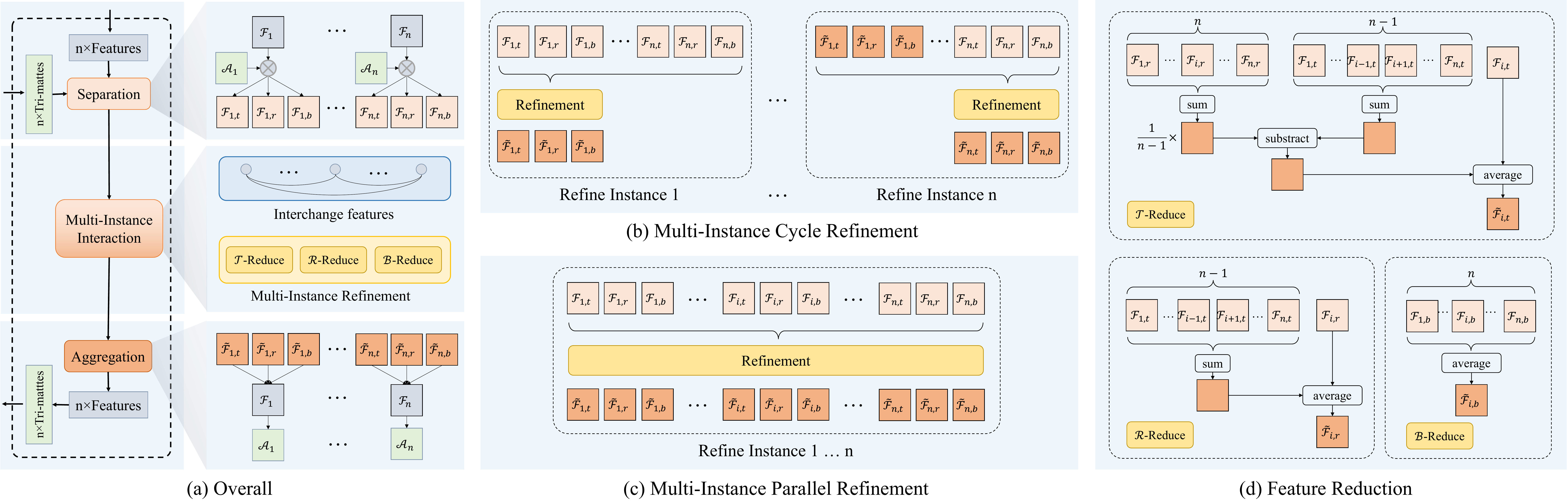}
    \vspace{-0.25in}
    \caption{(a) Structure of our multi-instance refinement module, where instances exchange information among each other to refine their features through a multi-instance interaction layer. Two representative multi-instance refinement strategies, i.e. (b) cycle refinement and (c) parallel refinement are proposed and discussed. Figure~(d) illustrates three feature reduction operations used in the two refinement ways.}
    \label{fig:CRH}
    \vspace{-0.2in}
\end{figure*}

\vspace{-2mm}
\paragraph{Mutual Information and Tri-mattes.}
To estimate target instance alpha matte $\alpha_i$, the other instances $j \not= i$ can be regarded as reference information. Note that we do not regard the other instances as part of background since they have different semantic representation. Therefore, we can re-formulate Equation~\ref{eq:im} into the following equation using {\em three} components, i.e., target instance $\mathcal{T}$, the other instances if any $\mathcal{R}$ (also named reference instances), and the background $\mathcal{B}$:
\begin{equation}\label{eq:im3}
I = \underbrace{\alpha_iL_i}_{\text{target $\mathcal{(T)}$}} +  \underbrace{\alpha_0L_0}_{\text{background $\mathcal{(B)}$}} + \underbrace{\sum_{j=1\ \text{and}\ j\ne i}^{n}\alpha_jL_j}_{\text{reference instances $\mathcal{(R)}$}}
\end{equation}
If we treat the component $\mathcal{R}$ as a new combined layer, Equation~\ref{eq:im3} is then simplified into a sparse representation as Equation~\ref{eq:im3_simple}, which considers the sparsity constraint:
\begin{equation}\label{eq:im3_simple}
I = \alpha_tL_t + \alpha_bL_b + \alpha_rL_r, \hspace{0.1in}
\textrm{s.t.}\ \alpha_t + \alpha_b + \alpha_r = 1
\end{equation}
where subscripts $t$,$r$,$b$ represent the three components $\mathcal{T}$, $\mathcal{R}$,$\mathcal{B}$ respectively. For a target instance, Equation~\ref{eq:im3_simple} implies that the alpha matte of each pixel can be correspondingly decomposed into three components,
$\alpha_t$, $\alpha_r$ and $\alpha_b$ (where one or two of them can be zero). These three components provide mutual information for one another, and they are collectively termed {\bf tri-mattes}.  

\subsection{Mutual Guidance Strategy}
Given an image, we first apply  MaskRCNN~\cite{MaskRCNN} to extract coarse masks $M$ for human instances. The challenge lies in turning the coarse mask into precise alpha matte for each instance. When only one instance exists, the task reduces into conventional human matting, which can be addressed by ~\cite{MG} or other matting techniques. 
To handle multiple instances, according to the above observations, we propose a novel mutual guidance strategy implemented using  a {\em tri-mask}. Tri-mask $\mathcal{M}$ is defined as the concatenation of $M_t$, $M_r$ and $M_b$, which respectively mask the region of $\mathcal{T}$, $\mathcal{R}$ and $\mathcal{B}$. For instance $i$, $M_{i,t}$, $M_{i,r}$ and $M_{i,b}$ are computed using the following tri-mask generation formulas,
\begin{align}
    M_{i,t} &= M_i , \hspace{0.2in} M_{i,r} = \bigcup\limits_{j=1\ \text{and} \ j\ne i}^{n} M_{j} \label{eq:mask_tr}\\
    M_{i,b} &= 1 - M_{i,t}\cup M_{i,r} \label{eq:mask_b}
\end{align}
Afterward, for each instance, we feed as input the concatenation of the image and its tri-mask into a matting branch for extracting its {\em tri-matte} $\mathcal{A}$, which is the concatenation of the alpha matte $\alpha_t$, $\alpha_r$ and $\alpha_b$. 
The matting branch is an encoder-decoder matting network adopting the same structure with the network used in ~\cite{MG}. After the matting branch, we extract the tri-mattes for all instances. To supervise $\mathcal{A}$, multi-instance constraints are employed which will be introduced in Section~\ref{sec:constraints}. 

Prior information in tri-mask provides comprehensive guidance for the model in pixel decomposition. On the one hand, the mutual exclusion among $M_t$, $M_r$ and $M_b$ guides the model to distinguish human instances from the background. On the other hand, the separation between $M_t$ and $M_r$ guides the model to differentiate instances. Subject to the constraint $\alpha_t + \alpha_r + \alpha_b = 1$, we force the model to learn a mutual exclusive decomposition in a contrastive manner.

\vspace{-0.1in}
\subsection{Multi-Instance Refinement}
Given $n$ instances, $n$ tri-mattes, i.e., $n$ triplets of $(\alpha_t, \alpha_r, \alpha_b)$ are derived via the aforementioned mutual guidance, 
which encourages intra-instance but not inter-intance consistencies, which
may lead to  misalignment among overlapping tri-mattes from different instances. We utilize 
such inter-instance inconsistencies to 
correct potential error of the estimated alpha mattes. Based on tri-mattes, we design a multi-instance refinement module (MIR), illustrated in Figure~\ref{fig:CRH} to further promote the quality of alpha mattes for all target instances. 

\vspace{0.05in}
\noindent\textbf{Overall Structure.} Our multi-instance refinement module comprises of three steps: separation, interaction and aggregation as shown in Figure~\ref{fig:CRH}-(a). For each instance, we use $\mathcal{F}_i$ to represent the feature from the final layer before the prediction head in the matting branch. Though $\mathcal{F}_i$ embodies the information for $\mathcal{T}$, $\mathcal{R}$ and $\mathcal{B}$, it is infeasible to perform individual operation on these three components. Thus, we use tri-matte to provide spatial attention so as to obtain the separate features for $\mathcal{T}$, $\mathcal{R}$ and $\mathcal{B}$. 
Specifically, multiplied by $\alpha_{i,t}$, $\alpha_{i,r}$ and $\alpha_{i,b}$, we obtain three features $\mathcal{F}_{i,t}$, $\mathcal{F}_{i,r}$ and $\mathcal{F}_{i,t}$, $i\in\{1,2,...,n\}$.

Separate representations for $\mathcal{T}$, $\mathcal{R}$ and $\mathcal{B}$ enable free communications and interactions to a large extent among instances. In the second step, a novel \textbf{multi-instance interaction} layer is proposed, in which each instance sends its features to other instances and receives the features from other instances. As the number of features varies with the number of instances, feature reduction operation is required to integrate these received features for refinement. 
Specifically, the refinement consists of three reduction operations, i.e., $\mathcal{T}$-reduce, $\mathcal{R}$-reduce, and $\mathcal{B}$-reduce, which are defined in Equation~\ref{eq:reduce_t}--\ref{eq:reduce_b} (Figure~\ref{fig:CRH}-(d)).
\begin{align}
\vspace{-0.1in}
    \Tilde{\mathcal{F}}_{i,t} &= \frac{1}{2}(\mathcal{F}_{i,t} + \frac{1}{n-1}\sum\limits_{j=1}^n\mathcal{F}_{j,r} - 
    \sum\limits_{j=1\ \text{and} \ j\ne i}^n\mathcal{F}_{j,t}) \label{eq:reduce_t}\\
    \Tilde{\mathcal{F}}_{i,r} &= \frac{1}{2}(\mathcal{F}_{i,r} +  \sum\limits_{j=1\ \text{and} \ j\ne i}^n\mathcal{F}_{j,t}) \label{eq:reduce_r}\\
    \Tilde{\mathcal{F}}_{i,b} &= \frac{1}{n}\sum\limits_{j=1}^n\mathcal{F}_{j,b} \label{eq:reduce_b}
    \vspace{-0.1in}
\end{align}
Equation (7)--(9) can be regarded as an averaging process. Such `averaging' can provide communication among instances obtained from each individual branch to alleviate uncertainty and stabilize the convergence. After the multi-instance interaction layer, we reunify $\tilde{\mathcal{F}}_{i,t}$, $\tilde{\mathcal{F}}_{i,r}$ and $\tilde{\mathcal{F}}_{i,t}$ to produce an enhanced feature for tri-matte estimation.

\vspace{0.05in}
\noindent{\textbf{Cycle versus Parallel Refinement.}} In the multi-instance interaction layer, after instances interchanging features information, there are numerous refinement possibilities since instances can refine their features concurrently or successively. Here, we discuss two representative refinement strategies in the multi-instance interaction layer, i.e. cycle refinement and parallel refinement shown in Figure~\ref{fig:CRH}-(b) and (c) respectively:
\vspace{-0.05in}
\begin{itemize}
    \item \textit{Cycle refinement}. Instances refine their features with the help of other features sequentially. For example, instance~$1$ first refines its feature and then sends its refined feature to all other instances. Next, instance~$2$ refines its features with the refined features from instance~$1$ and the unrefined features from the rest instances, and so on. Finally, instance $n$ refines its features based on the refined features from all the other instances.
    \vspace{-0.05in}
    \item \textit{Parallel refinement}. Instances refine their features with the help of other features simultaneously. All instances refine their features based on the unrefined features from the other instances.
\end{itemize}

Both refinement strategies are effective in utilizing multi-instance mutual information to alleviate the effect of outliers. Since cycle refinement is order-sensitive, parallel refinement is preferable in non-interactive applications. We adopt parallel refinement in this paper. More comparisons and implementation details can be found in the supplementary materials. 

\subsection{Multi-Instance Constraint}\label{sec:constraints}
Conventional matting losses, i.e., alpha loss and pyramid Laplacian loss, are still applicable in instance matting. Specially, we apply alpha loss and pyramid Laplacian loss for $\alpha_t$, $\alpha_r$ and $\alpha_b$ separately. Their summations are denoted by $L_{\alpha}$ and $L_{\text{lap}}$.

Alpha loss and pyramid Laplacian loss directly regularize the distance between the estimated alpha matte and the ground truth, not considering composition constraint and alpha constraint among multiple instances as well as the background. We adapt the composition loss to accommodate multi-instance composition constraint as Equation~\ref{eq:loss_comp},
\begin{equation}\label{eq:loss_comp}
    L_{mc} = ||\alpha_{t}F_{t} + \alpha_{r}F_{r} + \alpha_{b}F_{b} - I||_1
\end{equation}

In addition, we employ multi-instance alpha constraint on tri-matte as Equation~\ref{eq:loss_alpha} to reduce the solution space: 
\begin{equation}\label{eq:loss_alpha}
    L_{m\alpha} = ||\alpha_{t} + \alpha_{r} + \alpha_{b} - 1||_1
\end{equation}

Finally, the total loss is the summation of the aforementioned losses as Equation~\ref{eq:loss_total},
\begin{equation}\label{eq:loss_total}
    L = L_{\alpha} + L_{lap} + L_{mc} + L_{m\alpha}
\end{equation}

We apply the loss defined in Equation~\ref{eq:loss_total} for the tri-mattes from both the matting branch and the multi-instance refinement module. 

\vspace{-0.1in}
\section{Benchmark}
Existing benchmarks are  designed for  instance segmentation such as COCO dataset~\cite{COCO}, or matting  such as Composition-1K~\cite{DIM}, but not for instance matting. They cannot provide a comprehensive evaluation for instance matting. In this paper, we propose a human instance matting benchmark called HIM2K, which is composed of two subsets, synthetic image subset and natural image subset respectively containing 1,680 and 320 images.

\vspace{2mm}
\noindent{\textbf{Synthetic Subset.}} We collect a variety of human images and carefully extract the human foregrounds. Then, we randomly select 2--5 such foregrounds $F_i$, and iteratively composite them onto a non-human background image sampled from BG20K~\cite{animalmatting}  following Equation~\ref{eq:composition} below, where $I_0$ is the background image:
\begin{equation}\label{eq:composition}
    I_i = \alpha_i F_i + (1-\alpha_i) I_{i-1}, i\in\{1,...,n\}
\end{equation}
Expanding Equation~\ref{eq:composition} for each foreground object layer, a uniform formula can be derived as Equation~\ref{eq:composition_expand}:
\begin{equation}\label{eq:composition_expand}
    I_i = I_0\prod_{j=1}^{i}(1-\alpha_j) + \sum_{j=1}^{i}\alpha_jF_j\prod_{k=j}^{i}(1-\alpha_k)
\end{equation}
\begin{figure}[t]
    \centering
    \includegraphics[width=1.0\linewidth]{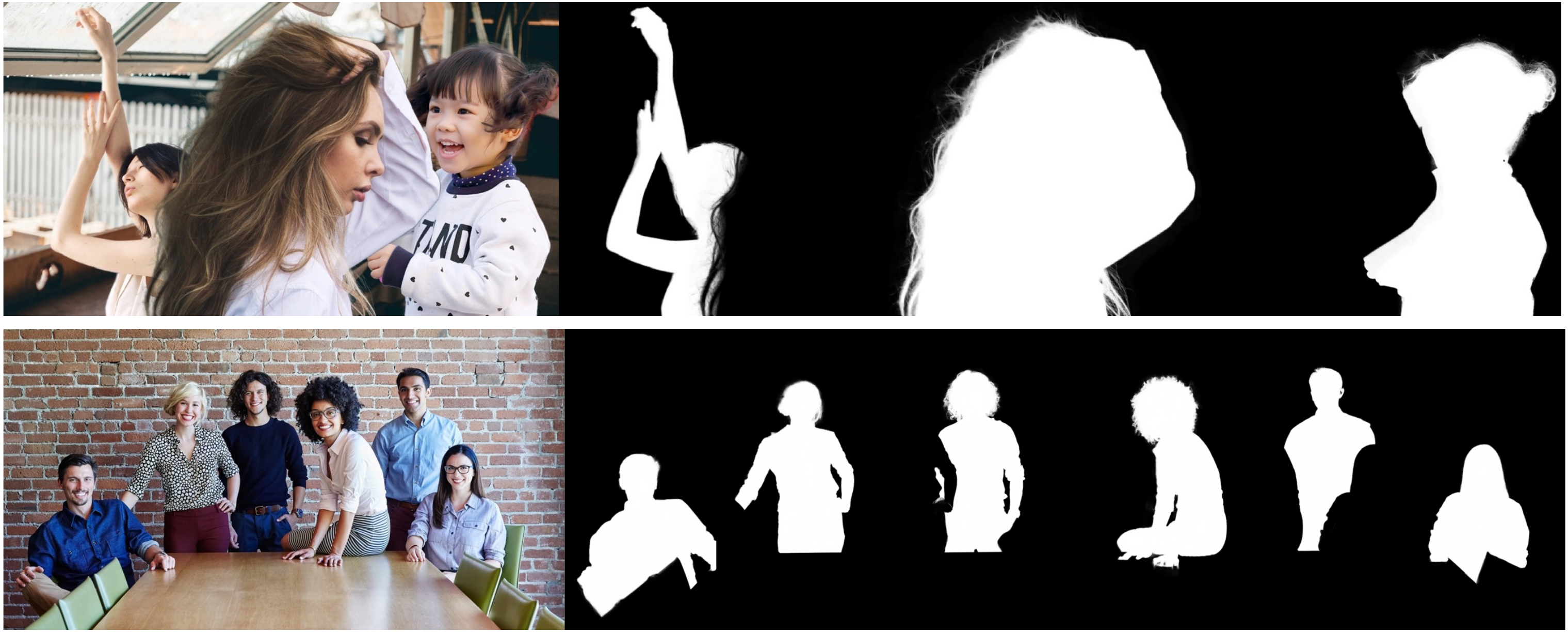}
    \vspace{-0.2in}
    \caption{HIM2K examples: top is synthetic and bottom is natural.}
    \label{fig:HIM2K}
    \vspace{-0.15in}
\end{figure}

If we use layer $L$ to represent a foreground image $F$ or background image $I_0$, Equation~\ref{eq:composition_expand} for the last iteration can be simplified as Equation~\ref{eq:composition_uniform} which is the same as Equation~\ref{eq:im}:
\begin{equation}\label{eq:composition_uniform}
    I = \sum_{i=0}^{n}\alpha_i^{'}L_i
\end{equation}
where $\alpha_i^{'}$ denotes the alpha matte of $i$-th layer $L_i$, the target to be estimated for instance $i$ when $i>0$.

\vspace{2mm}
\noindent{\textbf{Natural Subset.}} In light of the domain gap between  synthetic and real images, we construct a natural subset for fair evaluation. The natural subset consists of 320 images containing multiple human instances of a variety of poses and scenarios, with ground truth alpha matte obtained by manual labeling using Photoshop. Despite the possibly imperfect (still reasonably accurate) annotation, we found that more than 98\% of regions contain no more than 3 overlapping areas, which makes annotated ground truth trustworthy.  Evaluation on the natural subset can validate the effectiveness and stability of different methods on real-world photos. Figure~\ref{fig:HIM2K} shows examples from the two subsets.

\vspace{-0.1in}
\section{IMQ Metric}
\vspace{-0.1in}
In this section, we introduce a new metric for instance matting. Existing metrics are  designed for either matting or instance segmentation including semantic segmentation. Instance segmentation metrics, such as mask average precision (mask AP), are used for measuring the binary instance mask quality, and thus unsuitable for evaluating alpha matte with fractional values in transitional region. On the other hand, the most widely used matting metrics, namely, the four errors  MAD (or SAD), MSE, Gradient and Connectivity, measure alpha matte quality without instance awareness. The above limitations of existing metrics necessitate a new metric, which we call instance matting quality (IMQ).

\vspace{0.1in}
\noindent{\textbf{Instance Matting Quality.}}
IMQ measures  instance matte quality giving attention to both instance recognition quality and matting quality. 
Inspired by the panoptic quality~\cite{PQ}, IMQ is defined by Equation~\ref{eq:IMQ}: 
\begin{equation}\label{eq:IMQ}
    \text{IMQ} = \frac{\sum_{\alpha,\hat{\alpha}\in \mathit{TP}}S(\alpha,\hat{\alpha})}{|\mathit{TP}| + \frac{1}{2}|\mathit{FP}| + \frac{1}{2}|\mathit{FN}|}
\end{equation}
where $S$ is the similarity measurement function; $\mathit{TP}$, $\mathit{FP}$, and $\mathit{FN}$ are respectively the true positive, false positive and false negative sets; $\alpha$ and $\hat{\alpha}$ are the predicted and ground truth instance alpha matte. 
The computation of IMQ has two steps: {\em instance  matching} and {\em similarity measurement} as revealed in Equation~\ref{eq:IMQ}. 

\vspace{2mm}
\noindent{\textbf{Instance Matching.}}
To match the predicted instance mattes with ground-truth instance mattes, the matching criterion is intersection-over-union (IoU) between $\alpha$ and $\hat{\alpha}$. We first quantify each instance matte into a binary mask by applying $\alpha>0$ before computing IoU matrix. Based on the IoU matrix, we apply Hungarian matching~\cite{hungarian}, a greedy assignment strategy to achieve one-to-one assignment. All assigned predicted instance mattes are treated as TP candidates, where a candidate is assigned to TP if its IoU is above a threshold (0.5 is adopted in this paper). After settling the TP set, the FP set and FN set can be derived easily. 

\vspace{2mm}
\noindent{\textbf{Similarity Measurement.}} 
The similarity measurement criterion is defined as Equation~\ref{eq:score} below, where $w$ is a balance factor, and $\mathcal{E}$ is an error function, e.g., MSE,
\begin{equation}\label{eq:score}
    S(\alpha, \hat{\alpha})= 1-\min(w\mathcal{E}(\alpha, \hat{\alpha}), 1)
\end{equation}
We denote IMQ applying MSE error function to measure similarity as IMQ$_{\text{mse}}$. If we replace the error function $\mathcal{E}$ by MAD, Gradient and Connectivity, we respectively obtain  IMQ$_{\text{mad}}$, IMQ$_{\text{grad}}$ and IMQ$_{\text{conn}}$. 

\vspace{2mm}
\noindent\textbf{Analysis.} Similar to panoptic quality, IMQ can be decomposed into two components as Equation~\ref{eq:decomposition},
\begin{equation}\label{eq:decomposition}
\small{
    \text{IMQ} = \underbrace{\frac{\sum_{\alpha,\hat{\alpha}\in \mathit{TP}}\mathcal{S}(\alpha,\hat{\alpha})}{|\mathit{TP}|}}_{\text{Matting\ Quality\ (MQ)}}\underbrace{\frac{|\mathit{TP}|}{|\mathit{TP}| + \frac{1}{2}|\mathit{FP}| + \frac{1}{2}|\mathit{FN}|}}_{\text{Recognition\ Quality\ (RQ)}}
}
\end{equation}
RQ has a similar expression to $F_1$-score, a metric widely used in recognition tasks, while MQ measures the matting quality for TP set. Different from existing instance segmentation and matting metrics, collaboration of RQ and MQ provides a fair and comprehensive evaluation for instance matte quality.

\section{Experiments}
\vspace{-0.1in}
In this section, we introduce our synthetic training dataset, evaluation and ablation studies. More details about the implementation including the network structure, data augmentations and training schedule can be found in the supplementary materials.

\begin{figure*}
    \centering
    \includegraphics[width=1.0\linewidth]{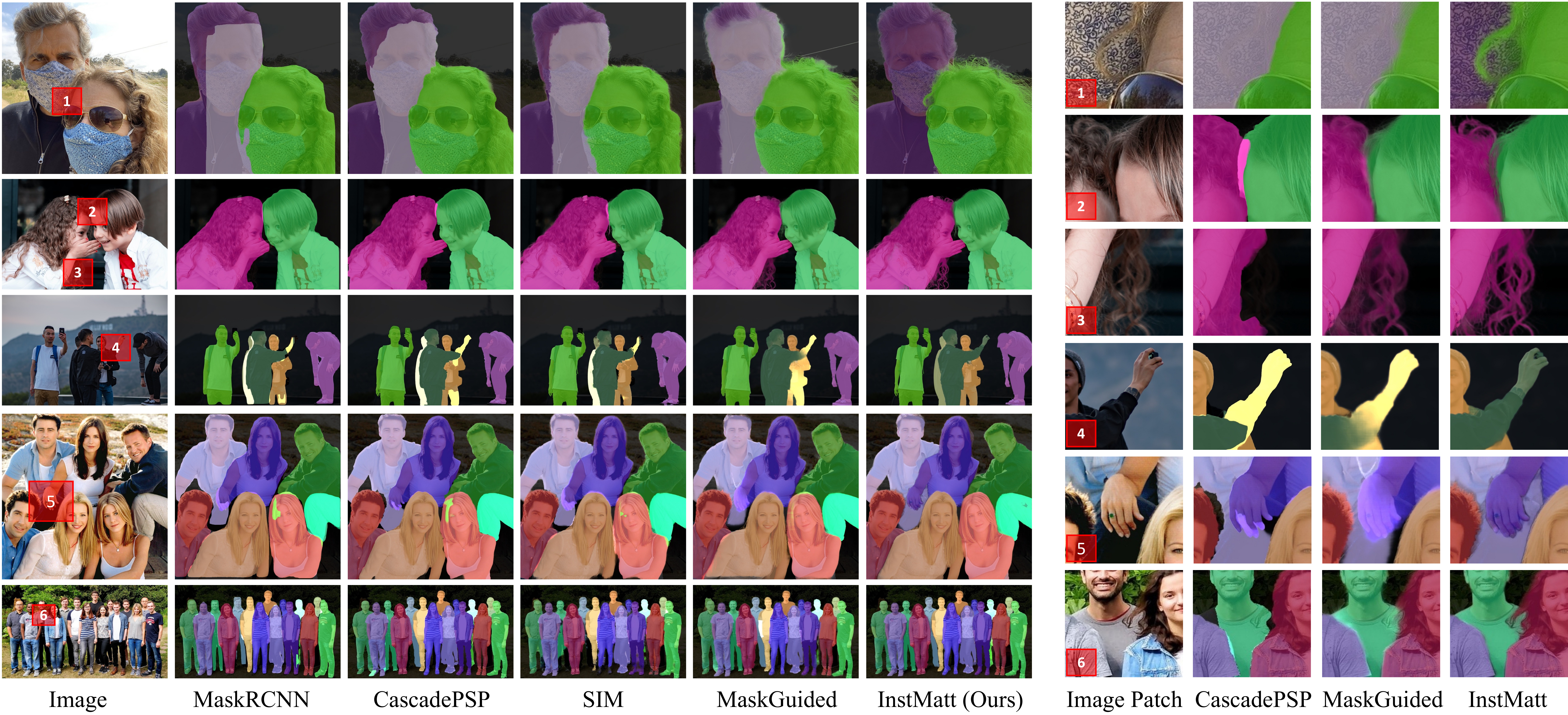}
    \vspace{-0.25in}
    \caption{Qualitative comparisons on real-world images. Right shows the of zoom-ins of patch 1--6.}
    \label{fig:qualitative}
    \vspace{-0.2in}
\end{figure*}

\subsection{Synthetic Training Dataset}
Since there is no off-the-shelf human instance matting training dataset, we construct our synthetic training dataset following~\cite{DIM}, by compositing human instances onto background images. Specifically, for the foreground, we collect 38,618 human instances with matting annotations from Adobe Image Matting dataset~\cite{DIM}, Distinctions-646~\cite{HAttNet} and self-collected dataset. For the background, we use non-human high-resolution images from~\cite{animalmatting,DVM}. 

\begin{figure}[t]
    \centering
    \includegraphics[width=1.0\linewidth]{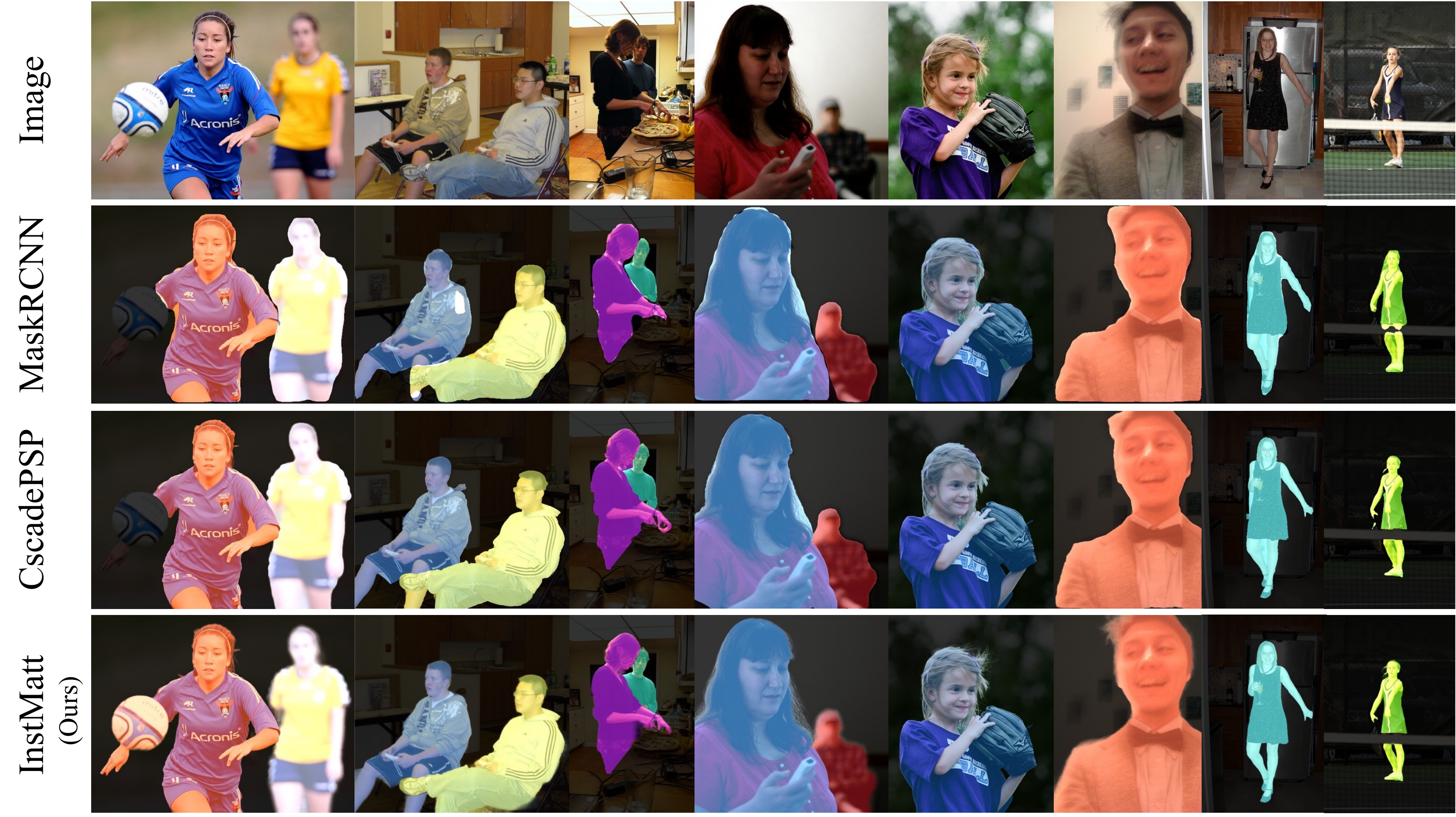}
    \vspace{-0.25in}
    \caption{Qualitative results on COCO dataset.}
    \label{fig:coco}
    \vspace{-0.2in}
\end{figure}

\begin{figure}[t]
    \centering
    \includegraphics[width=1.0\linewidth]{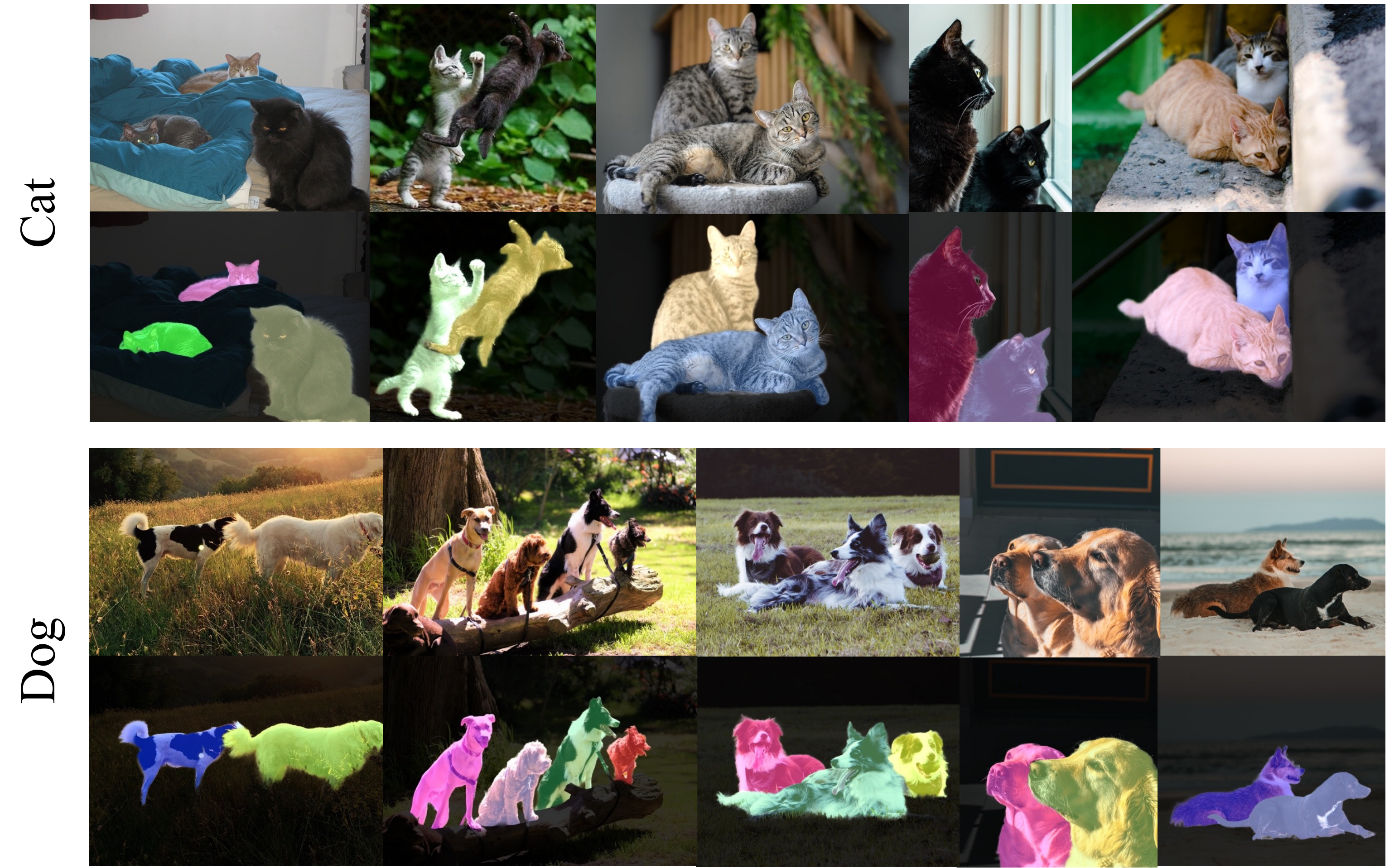}
    \vspace{-0.25in}
    \caption{Adaptation to other classes, i.e., cat and dog.}
    \label{fig:other_classes}
    \vspace{-0.3in}
\end{figure}

To produce a synthetic image, we randomly pick 2 to 5 instances from the foreground set, and composite them onto a background image. Random crop and zoom are applied on each foreground image. To avoid degenerate cases, such as a totally occluded instance, 
we  composite instances with a random gap or overlap within some reasonable range. The composition is an iterative procedure following Equation~\ref{eq:composition}. Finally, a total of 35,000 synthetic images with multiple instances are included in our training dataset.

\begin{table*}[t]
\footnotesize
    \centering
    \setlength{\tabcolsep}{1.2mm}{
    \begin{tabular}{l|c|c|c|c|c|c|c|c|c|c}
        \hline\hline
        \multicolumn{1}{c|}{\multirow{2}{*}{Method}} &
        \multicolumn{4}{c|}{\multirow{1}{*}{HIM2K (Synthetic Subset)}} &
        \multicolumn{4}{c|}{\multirow{1}{*}{HIM2K (Natural Subset)}} & 
        \multicolumn{2}{c}{\multirow{1}{*}{RWP636}} 
        \\
        \cline{2-4}\cline{4-11}
        \multicolumn{1}{c|}{} &
        \multicolumn{1}{c|}{\small{IMQ$_{\text{mad}}$}} &
        \multicolumn{1}{c|}{\small{IMQ$_{\text{mse}}$}} & 
        \multicolumn{1}{c|}{\small{IMQ$_{\text{grad}}$}} & 
        \multicolumn{1}{c|}{\small{IMQ$_{\text{conn}}$}} & 
        \multicolumn{1}{c|}{\small{IMQ$_{\text{mad}}$}} &
        \multicolumn{1}{c|}{\small{IMQ$_{\text{mse}}$}} & 
        \multicolumn{1}{c|}{\small{IMQ$_{\text{grad}}$}} & 
        \multicolumn{1}{c|}{\small{IMQ$_{\text{conn}}$}} 
        & 
        \multicolumn{1}{c|}{\small{IMQ$_{\text{mad}}$}} &
        \multicolumn{1}{c}{\small{IMQ$_{\text{mse}}$}} 
        \\
        \hline
        MaskRCNN~\cite{MaskRCNN} & 18.37 & 25.65 & 0.45 & 19.07 & 24.22 & 33.74 & 2.27 & 26.65 & 20.26 & 25.36
        \\
        MaskRCNN + CascadePSP~\cite{cascadepsp} & 40.85 & 51.64 & 29.59 & 43.37 & 64.58 & 74.66 & 60.02 & 67.20 & 42.20 & 52.91
        \\
        \hline
        MaskRCNN + GCA~\cite{GCA} & 37.76 & 51.56 & 38.33 & 39.90 & 45.72 & 61.40 & 44.77 & 48.81 & 33.87 & 46.47 \\
        MaskRCNN + SIM~\cite{SIM} & 43.02 & 52.90 & 40.63 & 44.29 & 54.43 & 66.67 & 49.56 & 58.12 & 34.66 & 46.60 \\
        MaskRCNN + FBA~\cite{FBA} & 36.01 & 51.44 & 37.86 & 38.81 & 34.81 & 48.32 & 36.29 & 37.23 & 35.00 & 47.54 \\
        MaskRCNN + MaskGuided~\cite{MG} & 51.67 & 67.08 & 53.03 & 55.38 & 
        57.98 & 71.12 & 66.53 & 60.86 & 30.64 & 53.16 \\
        \hline
        InstMatt (Ours) & \textbf{63.59} & \textbf{78.14} & \textbf{64.50} & \textbf{67.71} & \textbf{70.26} & \textbf{81.34} & \textbf{74.90} & \textbf{72.60} & \textbf{51.10} & \textbf{73.09} \\
        \hline\hline       
    \end{tabular}}
    \vspace{-0.1in}
    \caption{Quantitative comparisons on HIM2K and RWP636~\cite{MG}. The balance factor $w$ in Equation~\ref{eq:score} is set to 10. For IMQ$_{\text{mad}}$, IMQ$_{\text{mse}}$, IMQ$_{\text{grad}}$ and IMQ$_{\text{conn}}$, the higher, the better. Bold numbers indicate the best performance.}
    \label{tab:HHM2K}
    \vspace{-0.18in}
\end{table*}

\vspace{0.1in}
\subsection{Evaluation}
\vspace{-0.1in}
\noindent{\textbf{Human Instance Matting.}} We perform joint qualitative and quantitative evaluations on multiple datasets, including HIM2K, RWP636~\cite{MG}, SPD~\cite{supervisely}, COCO~\cite{COCO} dataset as well as more complex real-world images. 

HIM2K is the proposed benchmark for human instance matting. Since our method is the first work to address instance matting, we compare our method with instance segmentation methods~\cite{MaskRCNN, cascadepsp} and a straightforward extension on existing state-of-the-art matting methods~\cite{GCA, SIM, FBA, MG} based on the masks from MaskRCNN~\cite{MaskRCNN}. 
To validate the effectiveness of our method, we also conduct comparisons on a human matting benchmark, Real World Portrait 636 (RWP636), and a human segmentation dataset, Supervisely Person dataset (SPD). SPD consists of 5418 images with fine mask annotations. We split a subset comprising of 500 images from SPD as the testing dataset. Table~\ref{tab:HHM2K} and ~\ref{tab:SPD} tabulate the quantitative results on the three testing sets, showing our method achieves the state-of-the art performance. 

Figure~\ref{fig:qualitative} shows
qualitative comparisons on complex images, demonstrating that instance matting is capable of solving challenging cases with multiple and overlapping instances, which cannot be addressed by other existing instance segmentation or matting techniques. Note on the other hand while COCO is a widely used testing dataset in detection and segmentation tasks, the mask annotations are labeled by rough polygons thus making COCO inappropriate in quantitative results comparison for the instance matting task. Thus, we instead conduct qualitative comparisons on COCO dataset in Figure~\ref{fig:coco}. Compared with instance segmentation algorithms, our InstMatt framework is significantly better in handling complex matting scenarios at the instance level, such as defocus, motion, blurry or thin hairy structures. 

\begin{figure}[t]
    \centering
    \includegraphics[width=1.0\linewidth]{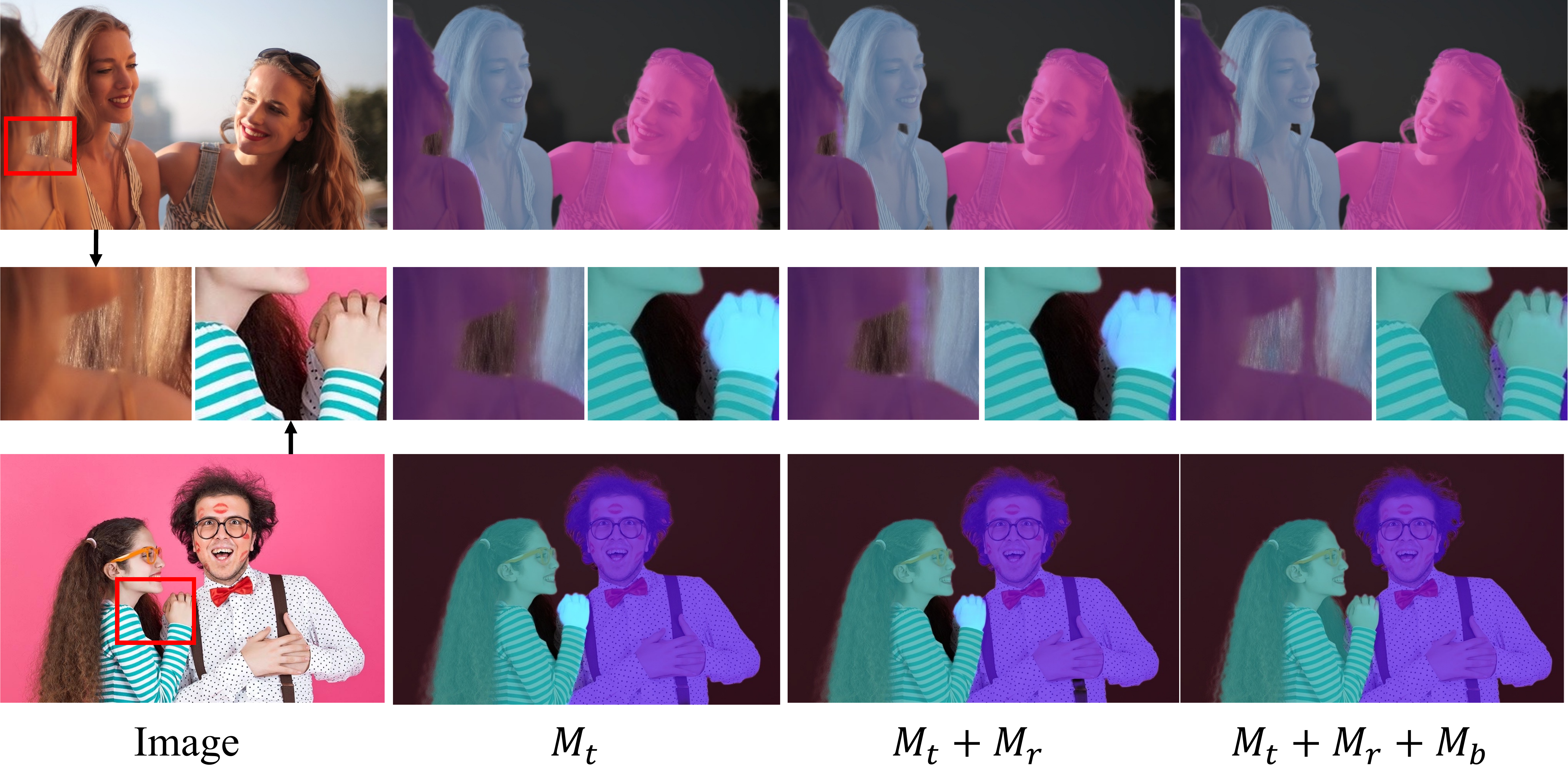}
    \vspace{-0.3in}
    \caption{Comparisons among different mask guidance settings. See in particular the zoom-ins showing the best results when all three components are enabled, where the blonde's hairs and the man's shoulder are clearly delineated.}
    \label{fig:tri_mask}
    \vspace{-0.1in}
\end{figure}

\begin{figure}[t]
    \centering
    \includegraphics[width=1.0\linewidth]{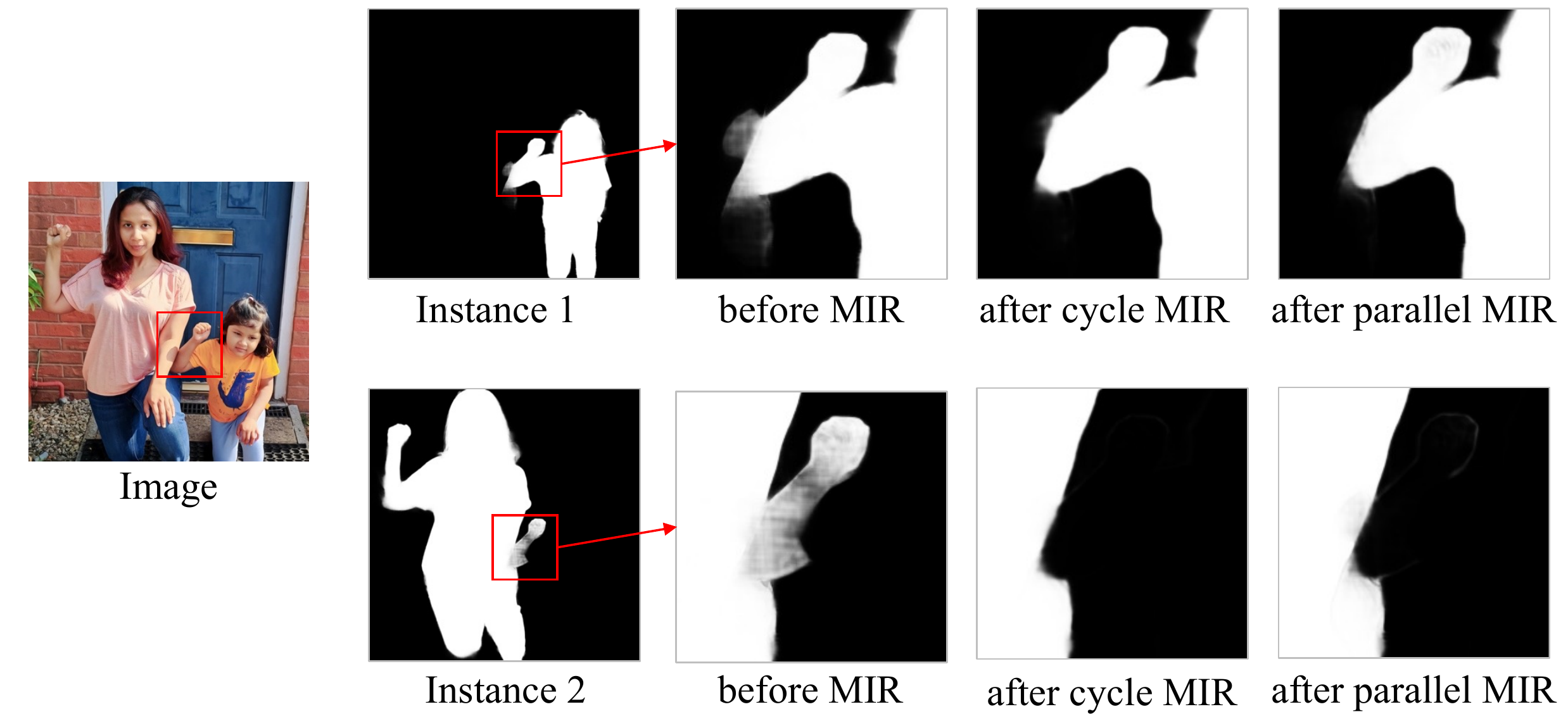}
    \vspace{-0.25in}
    \caption{Alpha matte before and after multi-instance refinement.}
    \label{fig:MIR}
    \vspace{-0.1in}
\end{figure}

\vspace{2mm}
\noindent{\textbf{Instance Matting Beyond Humans.}} This paper takes human instance matting as  our focused  contribution in instance matting. Notably,  our method, including mutual guidance, multi-instance refinement, multi-instance constraints, and the proposed instance matting metric IMQ as well can be also applied to instance matting on other semantic classes. We adapt our method on another two popular classes, i.e., cat and dog. Preliminary results shown in Figure~\ref{fig:other_classes} indicate that our method may generalize well to other semantic classes in instance matting.

\subsection{Ablation Study}
\noindent{\textbf{Tri-mask.}} The tri-mask provides mutual guidance for both instances versus background as well as instances versus instances. Table~\ref{tab:tri_mask} tabulates the results on models with different mask guidance settings. The tri-mask guides the model to assign each pixel partially to the target instance, other instances or the background. Notably, with tri-mask, some missing part due to occlusion are recovered as shown in the examples in Figure~\ref{fig:tri_mask}. The representation of the missing part is similar to that of the target instance, which cannot be ascribed to background or other instances due to the mutual exclusive supervision. 

\begin{table}[t]
\footnotesize
    \centering
    \setlength{\tabcolsep}{2.0mm}{
    \begin{tabular}{l|c|c}
    \hline\hline
         Method & IMQ$_{\text{mad}}$ & IMQ$_{\text{mse}}$\\
         \hline
         MaskRCNN~\cite{MaskRCNN} & 18.44 & 18.48 \\
         MaskRCNN + CascadePSP~\cite{cascadepsp} & 30.54 & 33.37 \\
         \hline
         InstMatt (Ours) & \textbf{30.67} & \textbf{39.56} \\
         \hline\hline
    \end{tabular}}
    \vspace{-0.1in}
    \caption{Quantitative results on SPD~\cite{supervisely}.}
    \label{tab:SPD}
    \vspace{-0.05in}
\end{table}

\begin{table}[t]
\footnotesize
    \centering
    \setlength{\tabcolsep}{2.5mm}{
    \begin{tabular}{c|c|c|c|c|c}
    \hline\hline
         $M_t$ & $M_r$ & $M_b$ & MIR & IMQ$_{\text{mad}}$ & IMQ$_{\text{mse}}$ \\
         \hline
         \cmark & \xmark & \xmark & \xmark & 57.98 & 71.12 \\
         \cmark & \cmark & \xmark & \xmark & 62.25 & 74.35 \\
         \cmark & \cmark & \cmark & \xmark & 69.40 & 79.74 \\
         \cmark & \cmark & \cmark & \cmark & \textbf{70.26} & \textbf{81.34} \\
         \hline\hline
    \end{tabular}}
    \vspace{-0.1in}
    \caption{Results on tri-mask and multi-instance refinement.}
    \label{tab:tri_mask}
    \vspace{-0.2in}
\end{table}

\vspace{1mm}
\noindent{\textbf{Multi-Instance Refinement.}} Multi-instance refinement aligns alpha matte predictions among multiple tri-mattes. Table~\ref{tab:tri_mask} shows that the IMQ$_{\text{mse}}$ of our model with and without multi-instance refinement module is 81.34 and 79.74, indicating an improvement from our multi-instance refinement. Figure~\ref{fig:MIR} further shows that
multi-instance refinement is helpful in erasing outliers due to the information synchronization among different instances.

\vspace{-0.05in}
\section{Conclusion}
\vspace{-0.05in}
In this paper, we propose a new task, instance matting with human instance matting as the first significant example by proposing a novel instance matting framework. Our InstMatt utilizes mutual exclusive guidance to guide the matting branch to extract alpha matte for each instance, which is followed by a multi-instance refinement module to synchronize information among co-occurring instances. InstMatt is capable of handling challenging cases with  multiple and overlapping instances, which can be adapted to other semantic class instance matting beyond human instances. We hope the proposed method, alongside with the new instance matting metric and the human instance matting benchmark, will encourage more future works.  

\newpage

\section*{A. Network Structure}
\vspace{-0.1in}
From the perspective of functionality, our InstMatt consists of two
steps, that is, instance recognition and mask refinement. We adopt MaskRCNN~\cite{MaskRCNN} with the backbone ResNet50~\cite{ResNet} as our instance recognition model since MaskRCNN is a conventional and also competitive approach in instance segmentation. We take the publicly released MaskRCNN pre-trained weight from Detectron2~\cite{detectron2} without further finetuning because this model is well-trained on the large-scale COCO~\cite{COCO} dataset containing rich scenarios which can be well generalized to other datasets.

The mask refinement step in InstMatt can be further divided into two modules, i.e., tri-mask guided matting branch and multi-instance refinement. We adopt the network used in MG~\cite{MG} as our matting branch. The network takes ResNet34~\cite{ResNet} as the backbone and applies three convolution blocks with a stride of $8$, $4$ and $1$ in the decoder respectively to reconstruct the features for tri-matte prediction. During inference stage, after extracting the instance masks, we obtain a tri-matte for each instance. Next, all the tri-mattes are sent to the multi-instance refinement for information synchronization. 

\noindent{\textbf{Multi-Instance Refinement.}} Considering the crowded cases with many instances, it is infeasible to perform multi-instance refinement on the whole image without out-of-memory problem when the memory storage is limited. Note that we only have to synchronize information on pixels which have information difference among all the instances and the background, that is, for a pixel $p$, 
\begin{equation}
    \sum_{i=1}^n\alpha_{p,i,t} + \alpha_{p,b} \ne 1
\end{equation}
For other pixels already satisfying the multi-instance alpha constraint, we may not gain much promotion from the information synchronization. 
Thus, we adopt the patch inference guided by an error map $E$, which is computed as follows,
\begin{equation}
    E = |\frac{1}{n}\sum_{i=1}^{n}\alpha_{i,b} + \sum_{i=1}^{n}\alpha_{i,t} - 1 | 
\end{equation}
We take the pixels with an error larger than 0.01 as the centers and correspondingly crop the patches of size $128\times128$ to perform multi-instance refinement. 

\begin{figure}[t]
    \centering
    \includegraphics[width=1.0\linewidth]{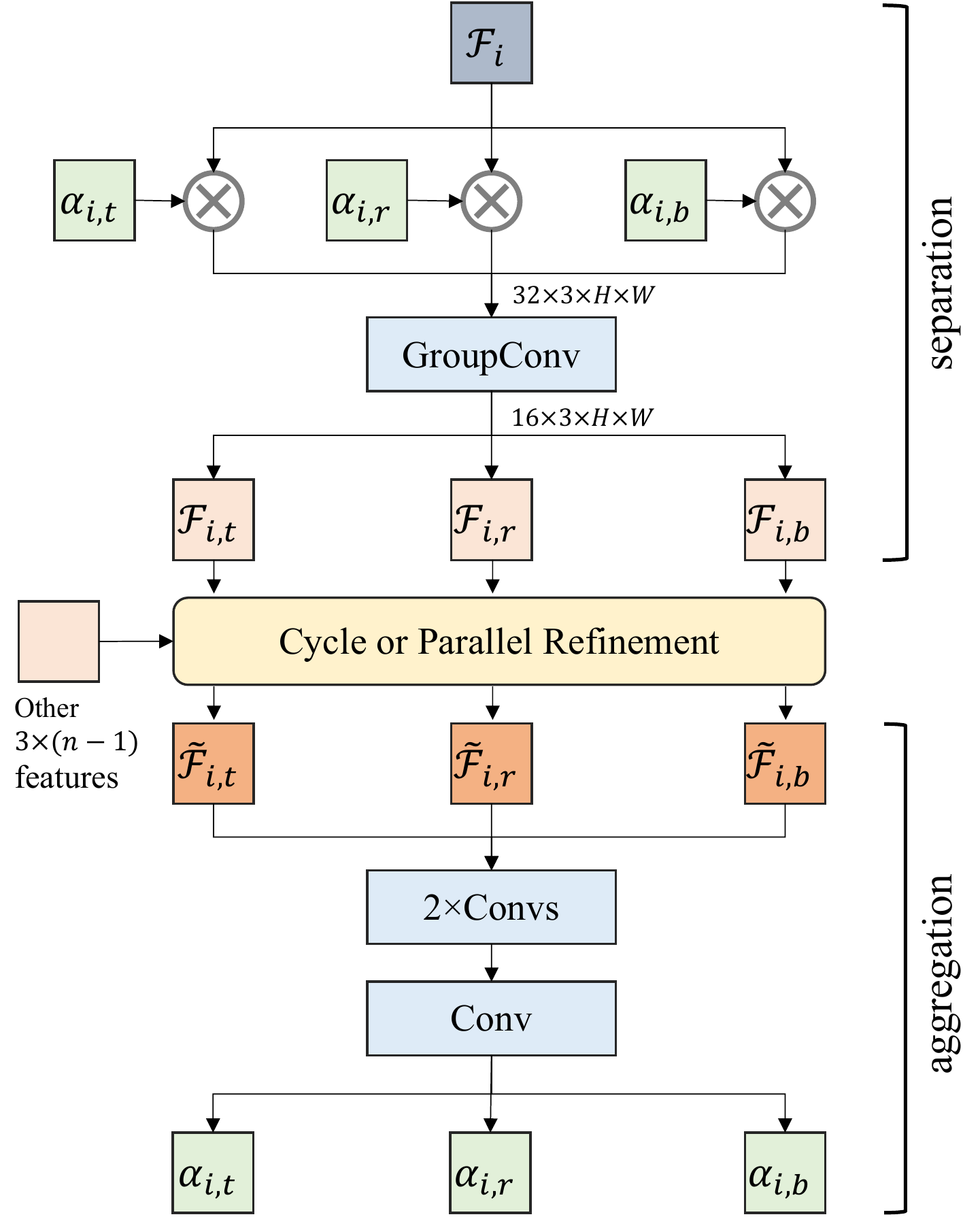}
    \caption{The structure of multi-instance refinement.}
    \label{fig:MIR_structure}
\end{figure}

In implementation, the multi-instance refinement contains 4 learnable layers as illustrated in Figure~\ref{fig:MIR_structure}. The four convolution layers all utilize $3\times3$ kernel. 

\begin{figure}[t]
    \centering
    \includegraphics[width=1.0\linewidth]{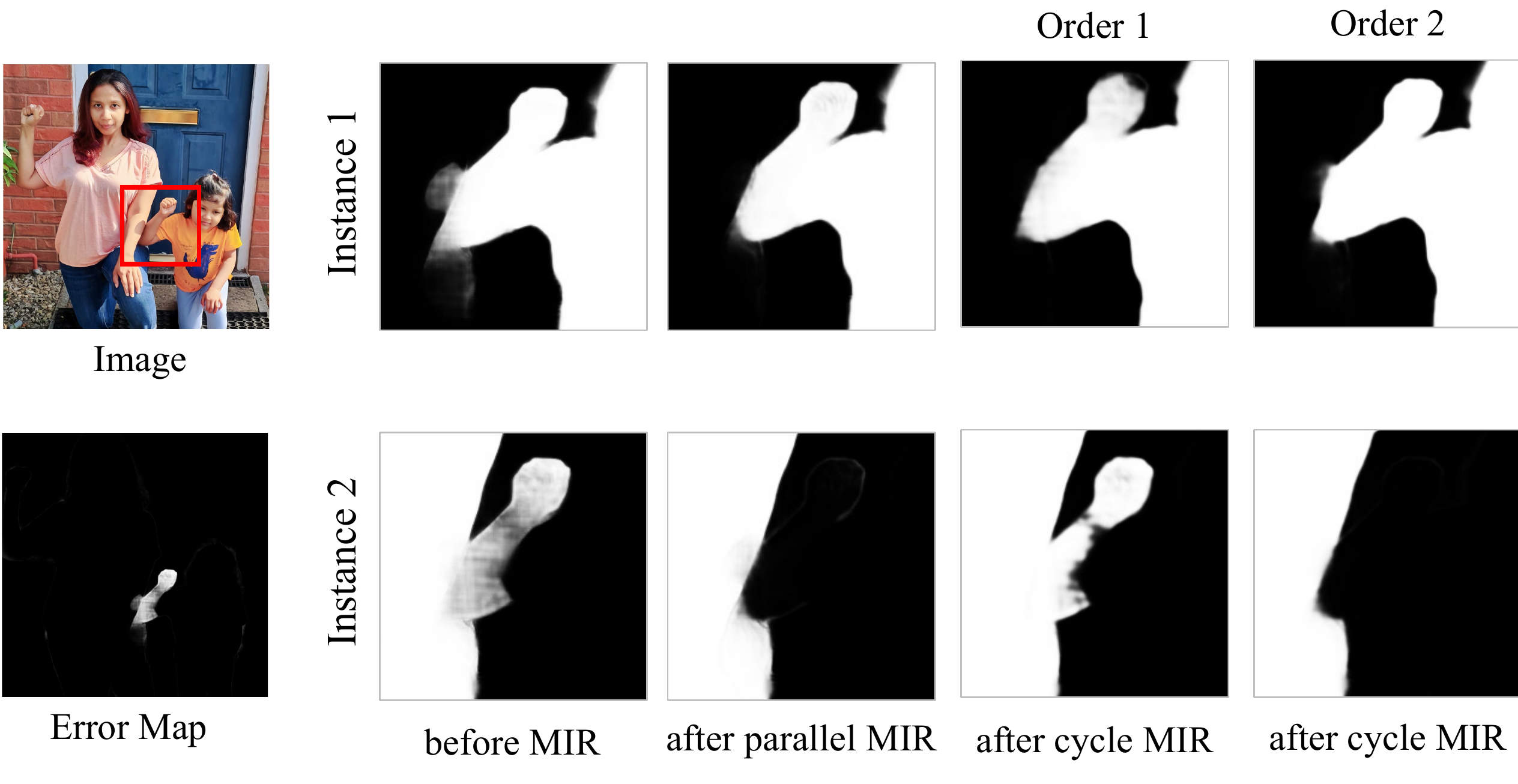}
    \caption{Cycle refinement versus parallel refinement.}
    \label{fig:cycle_parallel}
\end{figure}

\noindent{\textbf{Cycle versus Parallel Refinement.}} Cycle refinement is  order-sensitive, which is shown in the example in Figure~\ref{fig:cycle_parallel}. When adopting order 1 (instance 1, 2), the updated results get worse, while the outliers are perfectly removed when adopting order 2 (instance 2, 1). With user-supplied hints, cycle refinement is able to generate promising refined results. However, the instability makes the cycle refinement strategy inappropriate in non-interactive applications. On the contrary, the parallel refinement produces refined results not relevant to any order and thus shows stable performance.

\begin{figure}[t]
    \centering
    \includegraphics[width=1.0\linewidth]{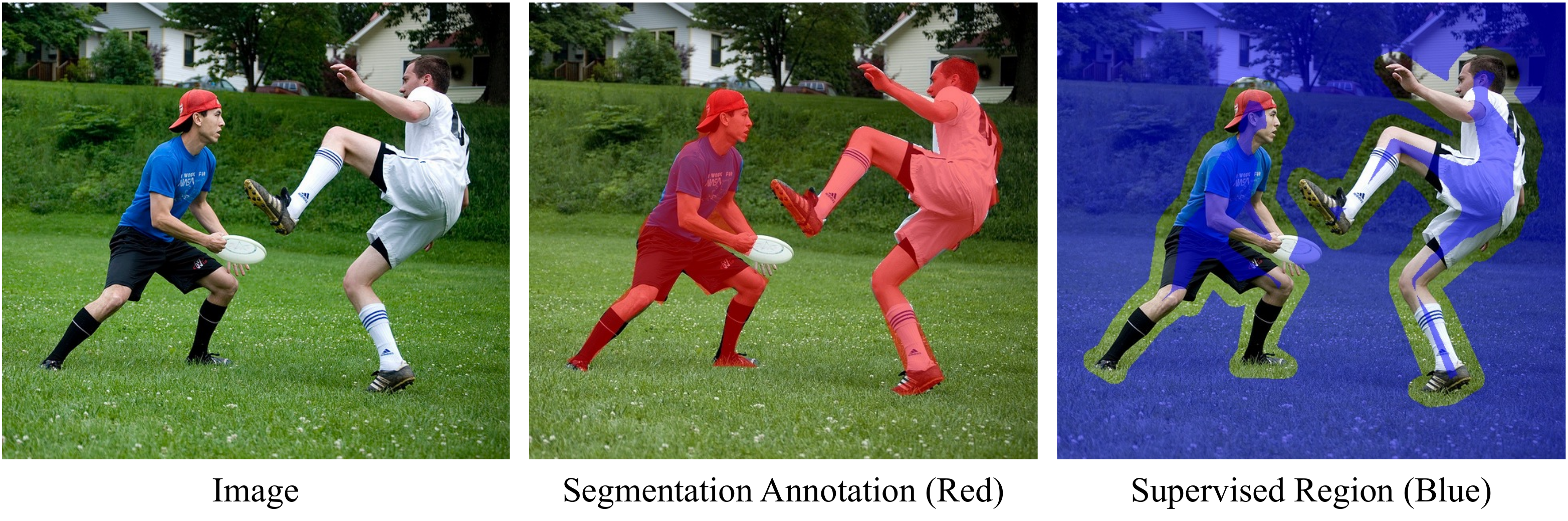}
    \caption{An example of partial supervision when adapting segmentation datasets to instance matting task. The segmentation mask annotations are drawn in red while the partial supervised region are highlighted in blue.}
    \label{fig:partial_supervision}
\end{figure}
 
\vspace{-0.1in}
\section*{B. Datasets}
During training, we use two datasets, one of which is the synthetic training dataset mentioned in the main paper Section 6.1. Since the synthetic training samples have a domain gap with natural images, we also include a natural set containing 41330 images selected from the COCO training set. However, it is non-trivial to adapt the COCO dataset to the instance matting task due to the lack of instance matting annotations. To tackle this issue, we adopt a \textbf{partial supervision} strategy to make the samples with segmentation annotations applicable in our task. 

\noindent{\textbf{Partial Supervision.}} See the example in Figure~\ref{fig:partial_supervision}. The instance segmentation annotations are labeled by polygons, which introduces noise along the boundary region. Thus, we respectively dilate and erode $k$ pixels along the boundary to generate a region shown in blue in Figure~\ref{fig:partial_supervision}, which are denoted as the supervised region while other pixels not masked are skipped. Such a partial supervision strategy allows us to make use of segmentation dataset without introducing noise. $k$ is set to 35 in training.

\section*{C. Augmentation}
To enrich the training dataset and avoid overfitting, various augmentation operations are adopted on the training samples. Besides random flip, random zoom, random shearing, as well as random crop, we  propose  \textbf{tri-mask augmentation} to improve the fault tolerance of the model, in particular, robustness against missing instances and imperfect masks.

\vspace{0.2in}
\noindent{\textbf{Missing Instance Tolerance}}. Let $M_r$ be the mask representing the union of all instances except for the target instance. Sometimes, the instance segmentation model is incapable of detecting all the instances. In this case, we only access a subset of the complete instance set to generate $M_r$. Therefore, the alpha constraint $\alpha_t + \alpha_r + \alpha_b = 1$ is no longer applicable. To avoid such a dilemma, we relax $M_r$ in the training stage to a subset of reference instances instead of the complete set.

\vspace{0.2in}
\noindent{\textbf{Mask Quality Tolerance.}} Equation 6 in the main paper mandates that $M_b$ is the complementary set of $M_t \cup M_r$. To avoid overfitting caused by such a strong constraint, we conduct dilation or erosion on the tri-mask after computing $M_t$, $M_r$ and $M_b$. In this way, the tri-mask may exhibit various gaps or overlaps among each other, thus introducing some uncertainty in training to better accommodate possible uneven quality of segmentation tasks.

\vspace{0.2in}
\noindent{\textbf{Tri-mask Augmentation.}} Due to the aforementioned two robustness considerations, we generate tri-masks in three steps, 1) instance mask generation, 2) instance separation, 3) mask perturbation. 

In the first step, for an image with $n$ instances, we adopt two ways to generate masks for these instances. For a subset of the $n$ instances, we obtain their masks from the instance segmentation model; for the rest of the instances, we generate their masks from a random truncation on the ground truth alpha matte. A hybrid of the two ways in instance generation increases the diversity of masks.

In the second step, we first randomly pick an instance $i$ as the target instance, then randomly choose a subset from the rest of $n-1$ instances to produce $M_r$. Finally we obtain $M_b$ by $1-M_t\cup M_r$. Such relaxation operation on $M_r$ make the model ascribe the pixels of those undetected instances to $\alpha_r$, rather than $\alpha_t$ or $\alpha_b$. 

In the last step, we randomly dilate or erode or perform a hybrid of dilation and erosion on the tri-mask with  kernel size in $[1,30]$. Such  perturbation on tri-mask further improves the fault tolerance of our model.

Note that the ground truth tri-matte for the tri-mask are generated without the relaxation or perturbation operation. 

Their generation follows Equation~\ref{eq:alpha_t}--\ref{eq:alpha_b}:
\begin{align}
    \alpha_{i,t} &= \alpha_i \label{eq:alpha_t} \\
    \alpha_{i,r} &= \sum\limits_{j=1\ \text{and} \ j\ne i}^{n}\alpha_j \label{eq:alpha_r} \\
    \alpha_{i,b} &= 1 - \alpha_{i,t} - \alpha_{i,r}\label{eq:alpha_b}
\end{align}

\begin{table}[t]
    \centering
    \setlength{\tabcolsep}{3.0mm}{
    \begin{tabular}{l|c|c}
    \hline\hline
         Method & IMQ$_{mad}$ & IMQ$_{mse}$ \\
         \hline
         without tri-mask aug. & 67.51 & 76.54 \\
         with tri-mask aug. & 69.40 & 79.74 \\
         \hline\hline
    \end{tabular}}
    \vspace{-0.1in}
    \caption{Ablation study on tri-mask augmentation.}
    \label{tab:tri_mask_aug}
    \vspace{-0.1in}
\end{table}
Without and with the tri-mask augmentation, the IMQ$_{mad}$ of our InstMatt is 67.51 and 69.40 as tabulated in Table~\ref{tab:tri_mask_aug}, showing a promotion benefiting from the tri-mask augmentation. 

\section*{D. Training Schedule}
Our training schedule consists of two steps:
\begin{enumerate}
    \item Train the matting branch on the synthetic and natural training datasets. The branch is initialized with ImageNet~\cite{ImageNet} pre-trained weight. We use a batch size of 16 in total on 4 GPU cards. Adam optimizer with $\beta_1=0.5$ and $\beta_2=0.999$ is adopted. The initial learning rate is set to $0.001$ and decays at a cosine learning rate~\cite{sgdr, fast_sgd}. The training lasts for 100,000 iterations with a warm-up of the first 5, 000 iterations. 
    \item After the matting branch is well-trained, we freeze the matting branch and train the multi-instance refinement module. In the second step, we use a batch size of 4 in total on 4 GPU cards. The initial learning rate is set to $0.0001$. The training lasts for 25,000 iterations with a warm-up of the first 1, 000 iterations. We keep the other hyper parameters the same as those in the first step.
\end{enumerate}

\begin{figure}[t]
    \centering
    \includegraphics[width=1.0\linewidth]{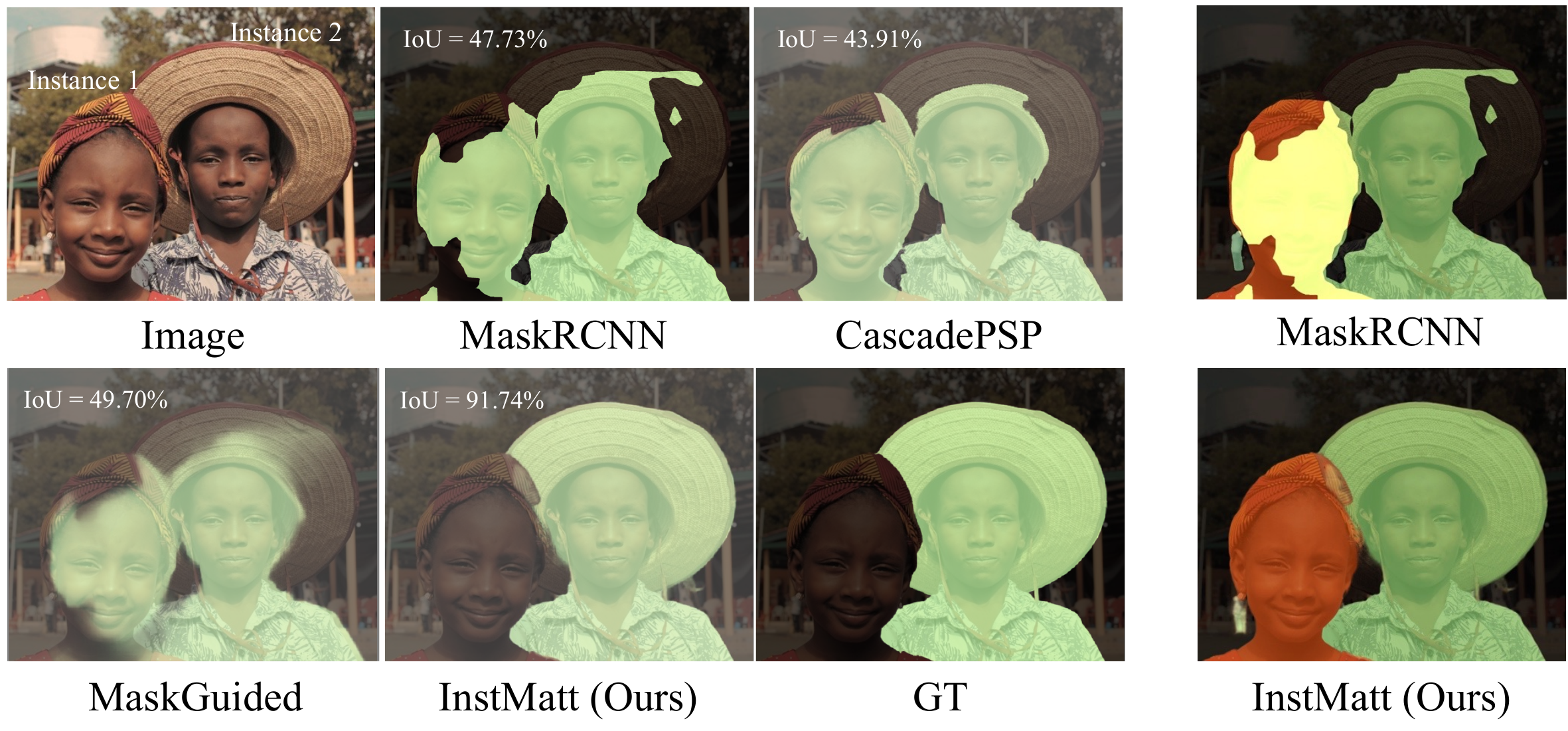}
    \vspace{-0.2in}
    \caption{IoU comparison. We show the predicted mask or alpha matte as from multiple methods as well as the ground truth for instance 2 in the left. CascadePSP and MaskGuided are both incapable of turning the wrong mask into a correct one while our InstMatt does. Right shows the results of two instances.}
    \label{fig:iou}
    \vspace{-0.1in}
\end{figure}

\begin{figure*}[t]
    \centering
    \includegraphics[width=1.0\linewidth]{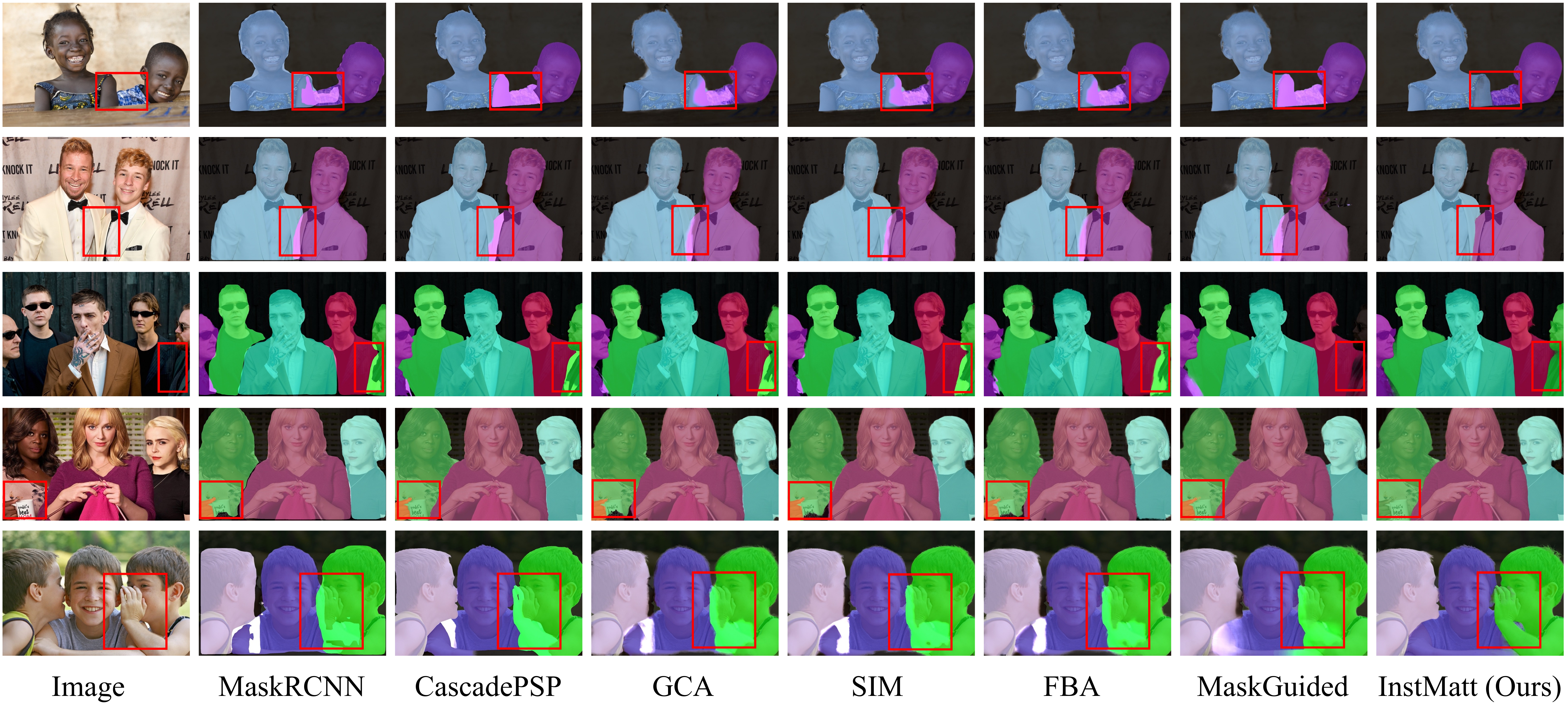}
    \vspace{-0.2in}
    \caption{Qualitative comparisons on images containing multiple instances in close proximity.}
    \label{fig:supp_close}
    \vspace{-0.1in}
\end{figure*}

\begin{figure*}[t]
    \centering
    \includegraphics[width=1.0\linewidth]{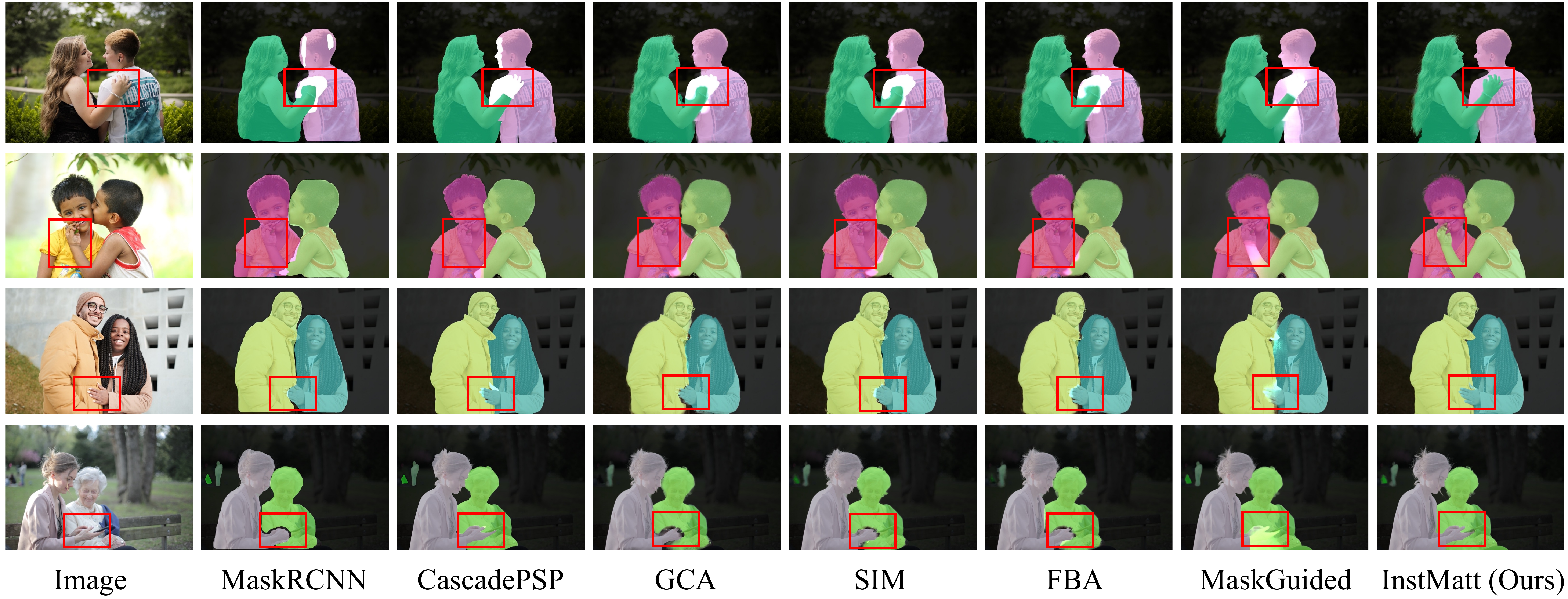}
    \vspace{-0.2in}
    \caption{Qualitative comparisons on images containing overlapping instances with long-range occlusion.}
    \label{fig:supp_overlap}
    \vspace{-0.1in}
\end{figure*}

\begin{figure*}[t]
    \centering
    \includegraphics[width=1.0\linewidth]{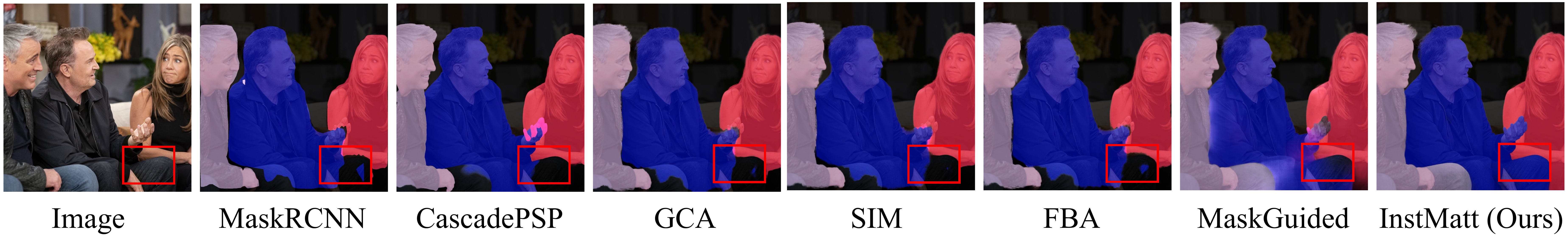}
    \vspace{-0.2in}
    \caption{Qualitative comparisons on a case with incomplete instance segmentation masks.}
    \label{fig:supp_missing}
    \vspace{-0.1in}
\end{figure*}

\begin{figure*}[t]
    \centering
    \includegraphics[width=1.0\linewidth]{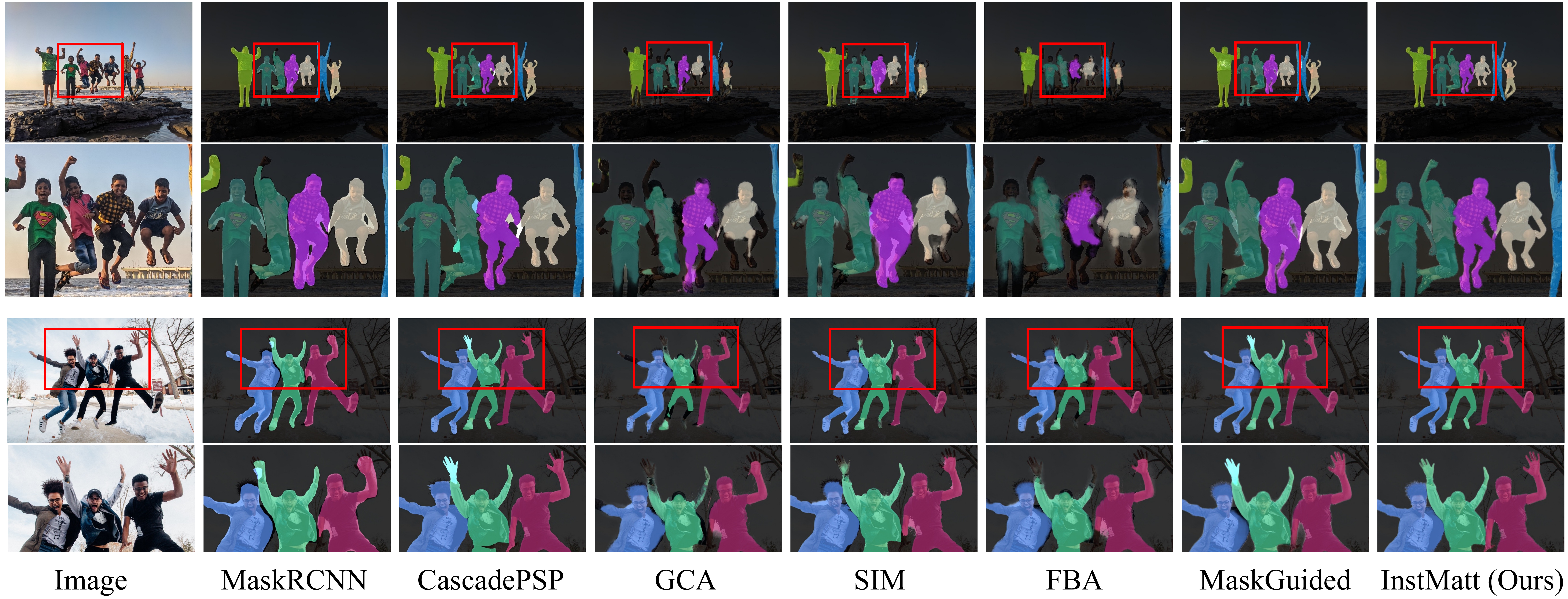}
        \vspace{-0.2in}
    \caption{Qualitative comparisons on images with small instances.}
    \label{fig:supp_small}
        \vspace{-0.1in}
\end{figure*}

\begin{table}[t]
    \centering
    \begin{tabular}{l|c|c|c}
    \hline\hline
         Method & IMQ$_{mad}$ & MQ$_{mad}$ & RQ \\
         \hline
         MaskRCNN~\cite{MaskRCNN} & 24.22 & 25.57 & 94.71 \\
         CascadePSP~\cite{cascadepsp} & 64.58 & 68.19 & 94.71 \\
         MaskGuided~\cite{MG} & 57.98 & 63.70 & 91.02 \\
         InstMatt (Ours) & \textbf{70.26} & \textbf{73.83} & \textbf{95.17} \\
         \hline\hline
    \end{tabular}
    \caption{MQ and RQ of MaskRCNN and our InstMatt.}
    \label{tab:IMQ_metric}
\end{table}

\section*{E. IMQ Metric}
\noindent\textbf{Computation.} We propose an IMQ metric to provide a comprehensive and unified evaluation of instance matte quality. The computation of IMQ can be divided into two steps, i.e., instance matching and similarity measurement. During matching instances, we first quantify the predicted and ground truth alpha mattes by applying $\alpha>0$ into binary mask to compute IoU matrix. Here, we use $0$ rather than other values as the threshold considering the semi-transparent/transparent objects usually composed of small alpha values. Other threshold will turn the alpha mattes of these objects into an incomplete binary mask which cannot cover the whole objects, thus leading to a wrong instance matching result. Quantification with 0 as threshold makes the IMQ metric applicable for not only human instance alpha mattes but also other semantic classes including transparent objects.

During similarity measurement, we adopt the widely used error functions in conventional matting task to evaluate the instance alpha matte from multiple dimensions. In implementation, we compute $\mathcal{E}(\alpha, \hat{\alpha})$ as follows,
\begin{align}
    \mathcal{E}(\alpha, \hat{\alpha}) &= \frac{1}{|\mathcal{P}|}\sum_{p\in \mathcal{P}}\mathcal{E}(\alpha_p, \hat{\alpha}_p) \\
    \mathcal{P} &= [\alpha>0]\cup [\hat{\alpha}>0]
\end{align}
We take the average upon the union of quantified $\alpha$ and $\hat{\alpha}$ instead of the whole image to avoid the overwhelming zero values from the large amount of background pixels especially for small instances.

\noindent\textbf{MQ and RQ.}
As mentioned in Section~5 in the main paper, IMQ can be decomposed of two components, RQ and MQ, measuring the instance recognition quality and the alpha matte quality of TP set respectively. We provide the RQ and MQ in Table~\ref{tab:IMQ_metric}. Compared to MaskRCNN, CascadePSP significantly promotes the instance matte quality among TP set, however, does not improves the RQ at all, demonstrating that CascadePSP cannot upgrade a low-quality mask which has an IoU below 0.5 with any ground truth instance mask into a high-quality instance mask due to the lack of instance awareness. 

On the contrary, besides refining the instance alpha matte along the boundary and the hairy regions among the TP set, our InstMatt is also capable of recognizing an instance and correspondingly extracting its alpha matte even though only a low-quality mask with misleading instance information is provided, such as the first example in Figure~5 in the main paper and the example in Figure~\ref{fig:iou}.

\section*{F. Experiment and Comparison}
\noindent{\textbf{Experiment Setting.}} We train our method InstMatt and MaskGuided~\cite{MG} on both the synthetic and natural datasets. For other methods including CascadePSP~\cite{cascadepsp}, GCA~\cite{GCA}, SIM~\cite{SIM} and FBA~\cite{FBA}, we use the released model from their official project website. To generate trimap for the trimap-based matting methods, we respectively dilate and erode the mask predicted from MaskRCNN~\cite{MaskRCNN} with a kernel size of 5 and then repeat the dilation and erosion operations for 10 times. 

\noindent{\textbf{Comparisons on HIM2K.}} Through our mutual guidance strategy in tandem with the multi-instance refinement module, our InstMatt shows superiority in various challenging cases. We provide more qualitative results for comparisons. 

Figure~\ref{fig:supp_close} shows the cases containing multiple instances next to each other closely. MaskRCNN is able to distinguish instances but produces overlapping instance masks, which cannot be addressed either by CascadePSP~\cite{cascadepsp} or a naive extension of existing matting models~\cite{GCA, SIM, FBA, MG}. Our InstMatt, however, can clearly separate the instances and generate non-overlapping instance alpha mattes.

Figure~\ref{fig:supp_overlap} shows the cases with occlusion. Under such cases, a part of one instance, usually a hand, or an arm, appears within the region of another instance and is far away from its own body. It is difficult to solve these cases for recognition tasks due to the limitation of receptive fields and the bottleneck of long-range feature propagation. As shown in Figure~\ref{fig:supp_overlap}, both instance segmentation models and matting models fall short of producing satisfactory results in these cases, while our InstMatt still produces promising results. Inter-instance mutual exclusive information guides the model to retrieve the remote pixels sharing the similar appearance with the body region instead of ascribing them to the other instances.

Figure~\ref{fig:supp_missing} compares the performance on a case with incomplete instance segmentation masks. Although CascadePSP or matting-based models are able to refine the mask along the boundary, they cannot recover a part of missing region due to the lack of instance awareness. Our InstMatt can find the lost region from the background giving the credit to the mutual exclusive guidance between the human instances and the background.

Figure~\ref{fig:supp_small} compares the performance on small instances. Compared to other methods, our InstMatt still shows stable performance on the instances of small or tiny scales, demonstrating the generalization ability and the fineness of our results.

\begin{figure}[t]
    \centering
    \includegraphics[width=1.0\linewidth]{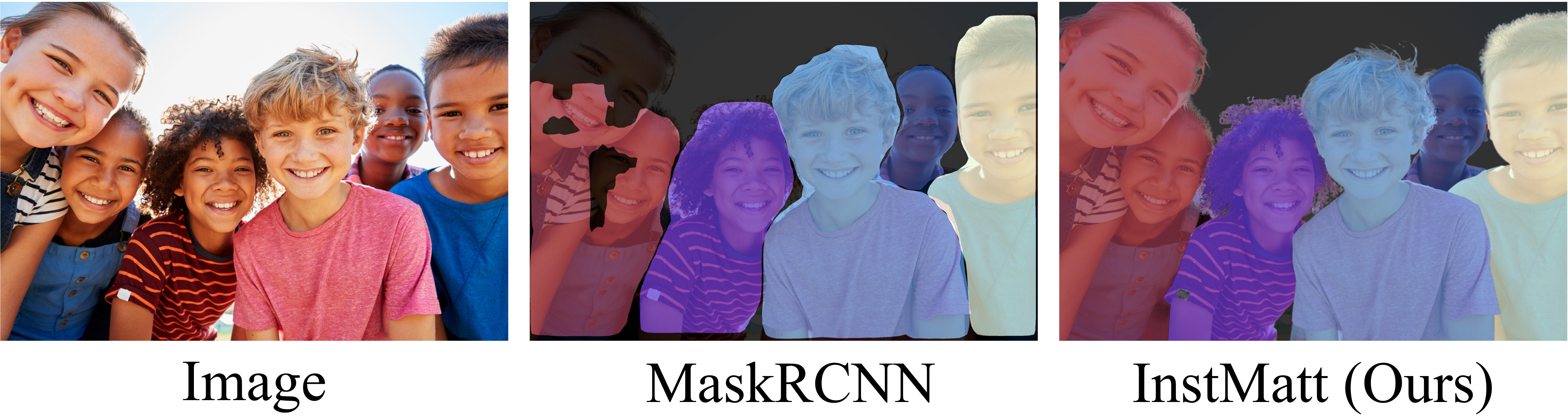}
    \vspace{-0.2in}
    \caption{An example explaining the limitation of our method.}
    \label{fig:limitation}
    \vspace{-0.1in}
\end{figure}

\vspace{-0.1in}
\section*{G. Limitation}
\vspace{-0.1in}
Under most cases, our method is capable of upgrading a low-quality instance mask into a high-quality alpha matte. However, sometimes the instance segmentation model cannot differentiate two largely overlapping instances as shown in Figure~\ref{fig:limitation}. Note that this example is different from the one in Figure~\ref{fig:iou}. In Figure~\ref{fig:iou}, MaskRCNN recognizes two instances although the mask of instance 2 covers a part of instance 1. Differently, in Figure~\ref{fig:limitation}, the instance segmentation model regards two left human instances as one and only predict a mask for the two left instances. In this case, our model can only refine the mask but cannot separate them due to the lack of sufficient guidance. 

\begin{figure}[t]
    \centering
    \includegraphics[width=1.0\linewidth]{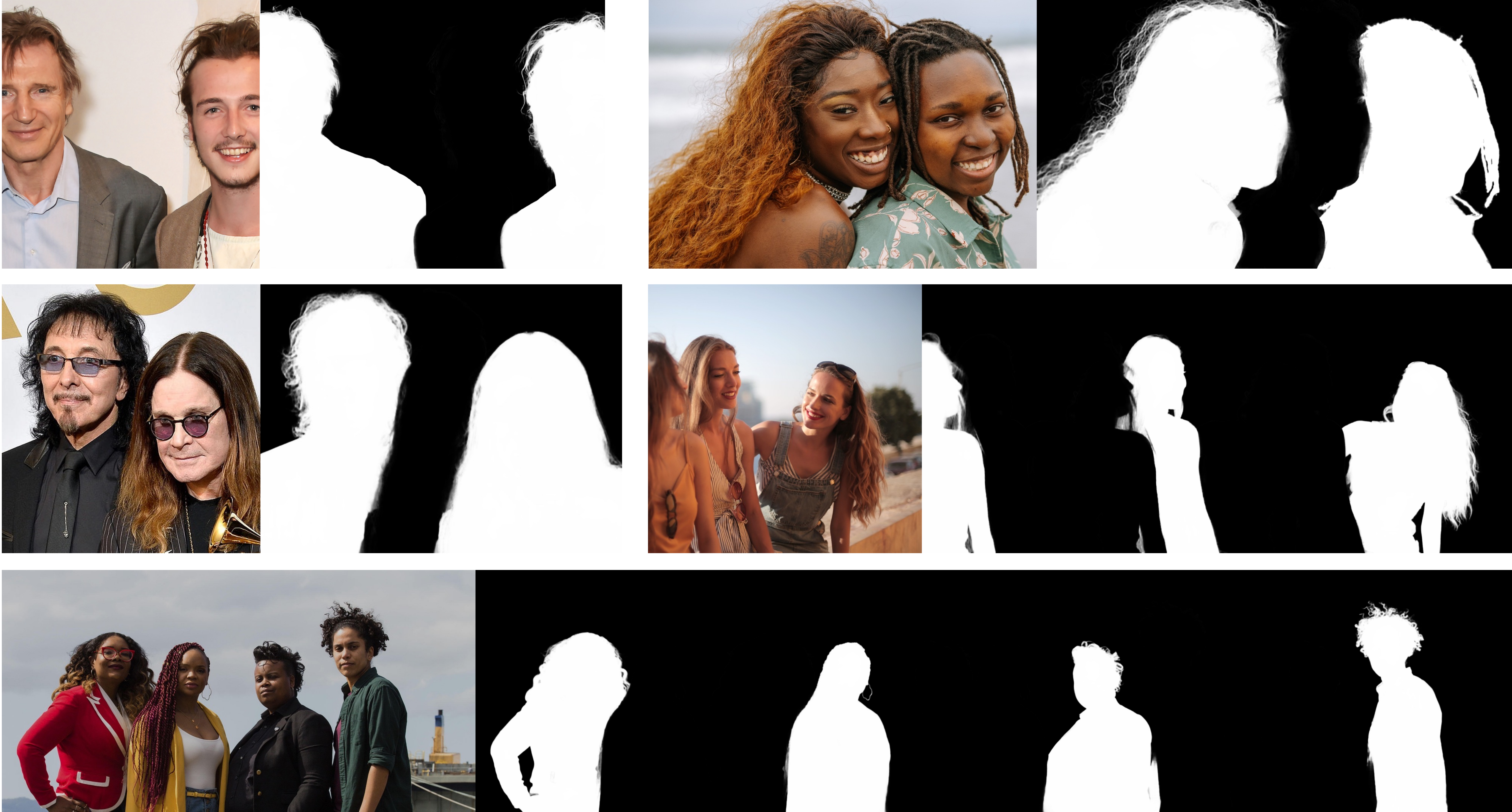}
    \vspace{-0.25in}
    \caption{Examples with hair details shown in alpha matte.}
    \label{fig:black_white}
    \vspace{-0.2in}
\end{figure}

\vspace{-0.05in}
\section*{H. More Qualitative Results}
\begin{figure*}[htp]
    \centering
    \includegraphics[width=1.0\linewidth]{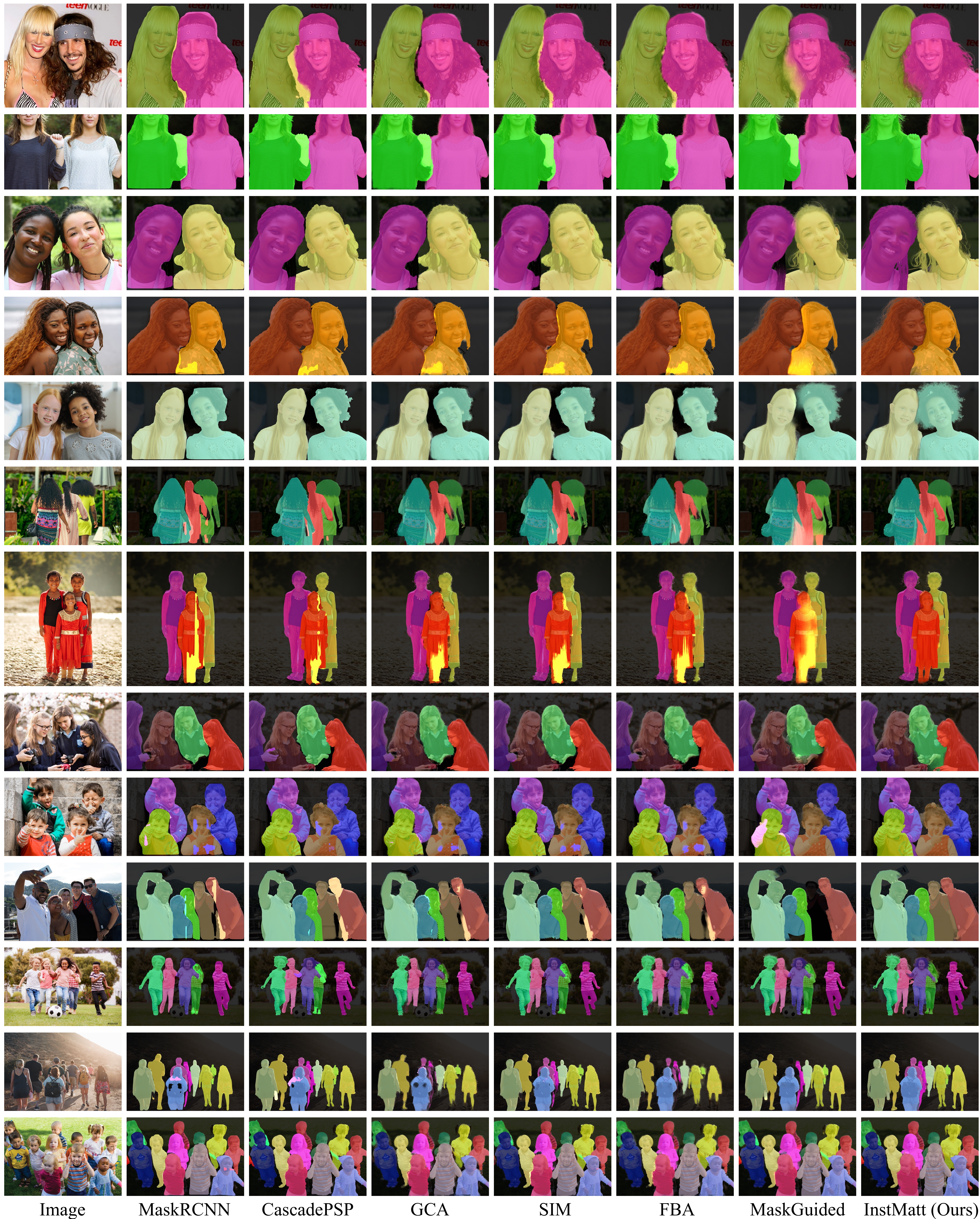}
    \caption{Qualitative comparisons on more complex cases with multiple overlapping or crowded cases.}
    \label{fig:supp_more}
\end{figure*}

\vspace{-0.05in}
Examples with hair details in alpha format are provided in Figure~\ref{fig:black_white}. More qualitative comparisons on images containing complex multiple overlapping or crowded cases are shown in Figure~\ref{fig:supp_more}.

{\small
\bibliographystyle{ieee_fullname}
\bibliography{main}
}

\end{document}